\documentclass[10pt,twocolumn,letterpaper]{article}

\usepackage{iccv}
\usepackage{times}
\usepackage{epsfig}
\usepackage{graphicx}
\usepackage{amsmath}
\usepackage{amssymb}

\usepackage{float}
\usepackage{subfig}
\usepackage{multirow}
\usepackage{array}
\usepackage{textcomp}
\newcolumntype{L}[1]{>{\raggedright\let\newline\\\arraybackslash\hspace{0pt}}m{#1}}
\newcolumntype{C}[1]{>{\centering\let\newline\\\arraybackslash\hspace{0pt}}m{#1}}
\newcolumntype{R}[1]{>{\raggedleft\let\newline\\\arraybackslash\hspace{0pt}}m{#1}}

\newcommand\Mark[1]{\textsuperscript{#1}}

\usepackage[pagebackref=true,breaklinks=true,colorlinks,bookmarks=false]{hyperref}
\iccvfinalcopy 

\def\iccvPaperID{9405} 

\begin{document}

\title{End-to-End Unsupervised Document Image Blind Denoising}

\author{
Mehrdad J Gangeh\Mark{1}\thanks{equal contribution}\qquad 
Marcin Plata\Mark{2}\footnotemark[1] \qquad  
Hamid R Motahari Nezhad\Mark{1} \qquad
Nigel P Duffy\Mark{1} \\
\Mark{1}Ernst \& Young (EY) LLP USA \qquad \Mark{2}EY GDS (CS) Poland Sp. z o.o. \\
{\tt\small \{Mehrdad.J.Gangeh,Hamid.Motahari,Nigel.P.Duffy\}@ey.com} \qquad 
{\tt\small Marcin.Plata@gds.ey.com}
}

\maketitle

\begin{abstract}
Removing noise from scanned pages is a vital step before their submission to optical character recognition (OCR) system. Most available image denoising methods are supervised where the pairs of noisy/clean pages are required. However, this assumption is rarely met in real settings. Besides, there is no single model that can remove various noise types from documents. Here, we propose a unified end-to-end unsupervised deep learning model, for the first time, that can effectively remove multiple types of noise, including salt \& pepper noise, blurred and/or faded text, as well as watermarks from documents at various levels of intensity. We demonstrate that the proposed model significantly improves the quality of scanned images and the OCR of the pages on several test datasets.
\end{abstract}

\section{Introduction}

Millions of electronic documents, such as contracts and invoices, are reviewed in the normal course of business in the enterprise. A large percentage of them are scanned documents containing various types of noise, including salt \& pepper (S\&P) noise, blurred or faded text, watermarks, etc. Noise in documents highly degrades the performance of the optical character recognition (OCR) and their subsequent digitization and analysis. The first step toward automating document analysis is to improve their quality using image processing techniques, such as image denoising and restoration. Most of the literature, places attention on removing noise from pictures ~\cite{Tian_review2020} (\eg, natural scenes) not text documents. However, these techniques may not be directly applicable due to very different nature of text documents.

In the image restoration problem, a degradation function and noise may both affect the quality of images~\cite{Gonzalez2006}. Examples are deblurring, defading, and inpainting. If in a special case, there is no degradation function, the problem would be a pure image denoising problem (\eg, S\&P noise removal). 


The current state-of-the-art (SOTA) solutions for image restoration problem are discriminative models based on convolutional neural networks (CNNs), auto-encoders and their variants such as REDNet (residual encoder decoder network)~\cite{RedNet2016,REDNET2019}, DnCNN (denoising convolutional neural networks)~\cite{DnCNN2017}, and RDN (residual dense network)~\cite{RDN2018,RDN2020}. These solutions can generally be formulated as follows:

\begin{equation}\label{eq:CNN}
    \underset{\mathbf{\theta}}{\mathrm{arg\,min}}\sum_j\sum_i\mathcal{L}\left (  f(\mathbf{x}_i^j;\mathbf{\theta)}=\hat{\mathbf{y}}_i^j, \mathbf{y}_i^j\right )
\end{equation}
where $\mathbf{x}_i^j$ refers to a noisy patch extracted from the noisy image $\mathbf{x}_i$, $\mathbf{y}_i^j$ and $\hat{\mathbf{y}}_i^j$ are the target and predicted clean patches, respectively, $f$ and $\theta$ refer to the CNN and its parameters. The main shortcoming of this approach, however, is the requirement for the availability of clean target images/documents, which is hard to address in real-world documents. In the literature, the noisy/clean pairs are usually prepared by adding some synthetic noise to clean images/documents. However, the synthetic noise does not completely model noise on real images/documents, and therefore, the performance of the network trained on these synthetic data is sub-optimal and highly degraded on real noisy images/documents~\cite{Kim2020,DBSN2020}.

To address the lack of noisy/clean pairs, noise-to-noise (N2N)~\cite{noise2noise2018}, noise-to-void (N2V)~\cite{noise2void2019}, and noise-to-self (N2S)~\cite{noise2self2019,self_supervised2019} training strategies have been proposed. However, these solutions are based on the assumption that the noise is additive zero-mean, and/or independent between pixels~\cite{noise2noise2018,noise2void2019,self_supervised2019}. This only covers a specific kind of denoising problem and, therefore, not directly applicable to general image restoration problems, including defading and deblurring. Furthermore, in N2N approach, at least two noisy instances of the same document page are needed, which are not readily available in real settings. 


There are several challenges in the design of an end-to-end solution for document image clean-up: 1) Noisy/clean pairs are not available, and therefore, the standard SOTA solutions based on discriminative models can not be employed. 2) There are various artifacts at different intensity levels (intra-class variation) in documents. 
3) We prefer to have a single model based on one architecture, and one training strategy, \ie, a \emph{unified} solution to address all noise/degradation problems (blind denoising/restoration) as opposed to individual models trained separately for each noise type. Training multiple models raises the problem of routing a document containing a specific artifact to the right model for image clean-up.

This paper addresses these challenges by introducing an end-to-end unsupervised image blind denoising algorithm that presents a single unified model to remove various noise types, without the requirement of paired noisy/clean pages. The main contributions are listed as follows:
\begin{enumerate}
\itemsep-0.4em
    \item We propose a novel unified architecture by integrating deep mixture of experts with a cycle-consistent GAN as the base network. We formulate a novel loss function for the proposed model.
    \item To the best of our knowledge, the designed unified model is the first that removes various artifacts, including noise (such as S\&P noise), and degradation (\eg, faded and blurred text, or watermark) at various intensity levels (image blind denoising). 
    \item We trained the model on actual noisy documents (not documents with synthetically added noise) without the requirement of noisy/clean images, evaluated it across several public and in-house document datasets, and demonstrated its excellent performance on real documents containing various artifacts. 
\end{enumerate}

\section{Related work}

\subsection{Discriminative methods}
Deep learning techniques for image denoising were pioneered by discriminative models and CNN/autoencoder architectures. 
Dong \etal~\cite{Dong2015} proposed one of the earliest CNN-based models for image denoising in the application of compression artifact reduction with several stacked convolutional layers. Ever since, various architectures and modifications of CNNs
have been proposed to improve image denoising, including skip symmetric connections in a model named residual encoder-decoder (REDNet)~\cite{RedNet2016,REDNET2019}, denoising CNN (DnCNN) employing batch normalization and residual learning~\cite{DnCNN2017}, residual dense networks (RDNs)~\cite{RDN2018,RDN2020}, wavelet CNNs~\cite{Wavelet-CNN2018}, feature attention~\cite{Anwar2019}, and dual residual networks (DRNs)~\cite{DualResidualNetworks2019}. 
For example, using DRNs~\cite{DualResidualNetworks2019}, the authors redesigned one network for each specific denoising task, including additive Gaussian noise removal, deblurring, dehazing, and raindrop removal. The main issue with discriminative approaches is that they need noisy/clean pairs.

\subsection{Generative adversarial methods}
Approaches based on generative adversarial networks (GANs)~\cite{Goodfellow2014,wGAN2017} are recently used to alleviate the requirement for noisy/clean pairs in several different ways. \Eg, Chen \etal~\cite{Chen2018} proposed to estimate the noise distribution from smooth parts of noisy patches using a GAN and to generate noise samples. They composed pairs of noisy/clean images by adding the estimated noise samples to clean patches, which are successively used to train a CNN for denoising. They claim that the estimated noise by the GAN is more realistic than the synthetic noise usually added in discriminative approaches. 
Cha \etal~\cite{GAN2GAN2020} went one step further and estimated noisy patches (instead of noise samples) using a GAN. This alleviated the requirement to have clean images. Having two different noisy instances of the same patch/image, they used N2N training strategy~\cite{noise2noise2018} to train a CNN for image denoising. Both previous approaches are merely applicable to situations where the noise is zero-mean, additive, and independent of the clean image, which are strong requirements that make them applicable to only a subset of, but not general image restoration problems. 

\subsection{Image denoising in documents}
Discriminative approaches for document image clean-up include a CNN-based approach for deblurring~\cite{BMVC2015}, a U-net~\cite{U-net2015} based approach replacing the skip connections between the encoder and decoder blocks with alternating convolutional and recurrent layers for efﬁcient feature extraction~\cite{SVDocNet2019}, a two-stage CNN-based approach where the first stage is to classify the type of deblurring and the second stage to remove it~\cite{Jiao2017}, and conditional GANs (cGANs)~\cite{DeepReader2018,DE-GAN2020}, which is a supervised image-to-image translation approach~\cite{cGAN2017}. DE-GAN~\cite{DE-GAN2020}, particularly, is recently proposed based on cGANs with a modified loss function with promising results on binarization, deblurring, and watermark removal. However, all these methods, including DE-GAN need noisy/clean pairs generated by adding the corruption to the clean pages/patches and train one model per artifact type. Sharma \etal~\cite{Sharma2018} proposed document image cleansing based on cycle-consistent GANs~\cite{Zhu2017}. This approach does not require noisy/clean pairs, however, they trained the model using these pairs~\cite{Sharma2018}. Also, they have trained one model for each noise/artifact type, whereas we have trained one single model for all types of artifacts.  

\section{Methods}

\subsection{Problem statement}


Here, we express the generic problem of document image clean-up in two statements: 

\noindent \textbf{Problem statement 1} - In real settings, there are many clean documents, which are unpaired to available noisy pages. The main question is whether we can take advantage of these clean documents in our solution. We can formulate the problem as follows: having two unpaired sets of documents, one set consisting of noisy documents ($\mathbf{X}$) and the other a collection of clean documents ($\mathbf{Y}$), and knowing that these two sets are unpaired, can we transform one set to have the style of the other? This problem can be formulated as an unsupervised image-to-image translation. Recently, there have been several proposals for solving this problem and one of the most prominent ones is based on cycle-consistent GANs (or cycle-GANs in short)~\cite{Zhu2017}. In fact, in this solution, what is more relevant to our problem is to transform the style of noisy documents to clean documents such that we remove/restore noise/degradation from these documents while preserving their text contents. 

\noindent  \textbf{Problem statement 2} - The noisy documents $\mathbf{X}$ may contain several different noise types. 
Our primary goal is to design a \emph{single model} that can tackle all these noise types. In our solution, we propose to integrate deep mixture of experts into the cycle-GAN. 


\subsection{Cycle-consistent GANs}\label{subsec:cycle-GAN}
Suppose there are two sets of unpaired document images $\{\mathbf{x}_i\}_{i=1}^N$ and $\{\mathbf{y}_j\}_{j=1}^M$ taken from two domains of noisy $\mathcal{X}$ and clean $\mathcal{Y}$ images, respectively. A cycle-GAN~\cite{Zhu2017} consists of two generators: forward ($H$) and backward ($F$) generators, and two adversarial discriminators $D_{\mathcal{Y}}$ and $D_{\mathcal{X}}$. The generators transform the data from one domain to another, \ie, $H: \mathcal{X}\rightarrow\mathcal{Y}$ and $F: \mathcal{Y}\rightarrow\mathcal{X}$. The adversarial discriminators aim to differentiate between the outputs of generators and the real data, \ie, $D_{\mathcal{Y}}$ aims to discriminate between $H(\mathbf{x})$ and $\mathbf{y}$, whereas $D_{\mathcal{X}}$ tries to distinguish between $F(\mathbf{y})$ and $\mathbf{x}$. The objective function in cycle-GAN is based on two losses: the GAN loss that transforms the image style from one domain to another, and the cycle-consistency loss that preserves the contents of the image. 

\subsection{Deep mixture of experts}
In order to address the problem statement 2, \ie, designing a single unified model that removes different types of noise from documents, we propose using a mixture of experts model. A na\"ive approach is to combine individual trained cycle-GANs (for each noise type) with an ensemble learning on top (Figure~\ref{fig:convMoE_cycleGAN}). This model-level combination of experts results in a very complex model that needs as many cycle-GANs as the number of noise types. 

Here, we propose an alternative solution with much less complexity based on deep mixture of experts (deep MoEs)~\cite{deepMoE2020}. Deep MoE operates within a single model and treats each channel (of a CNN, for example) as an expert. It extends the standard single layer MoE model to multiple layers within a single CNN. Wang \etal~\cite{deepMoE2020} derived the equivalence between gated channels in a convolution layer and the classic mixture of experts. A deep MoE consists of three main components: 1) base convolutional network, 2) shallow embedding network, and 3) multi-headed sparse gating network (refer to Figure~1 in~\cite{deepMoE2020} for the architecture of deep MoEs). The optimization function for the deep MoEs is based on minimizing three losses: 1) the loss for the base CNN (\eg, cross-entropy loss), 2) the cross-entropy loss for the shallow embedding network, and 3) the $\ell_1$ loss for the gated networks.  

\subsection{The proposed architecture: integrated cycle-GAN and deep MoE}

\begin{figure*}[!tbh]
    \centering
    \subfloat[]{\label{fig:convMoE_cycleGAN} 
    \includegraphics[width=.28\textwidth]{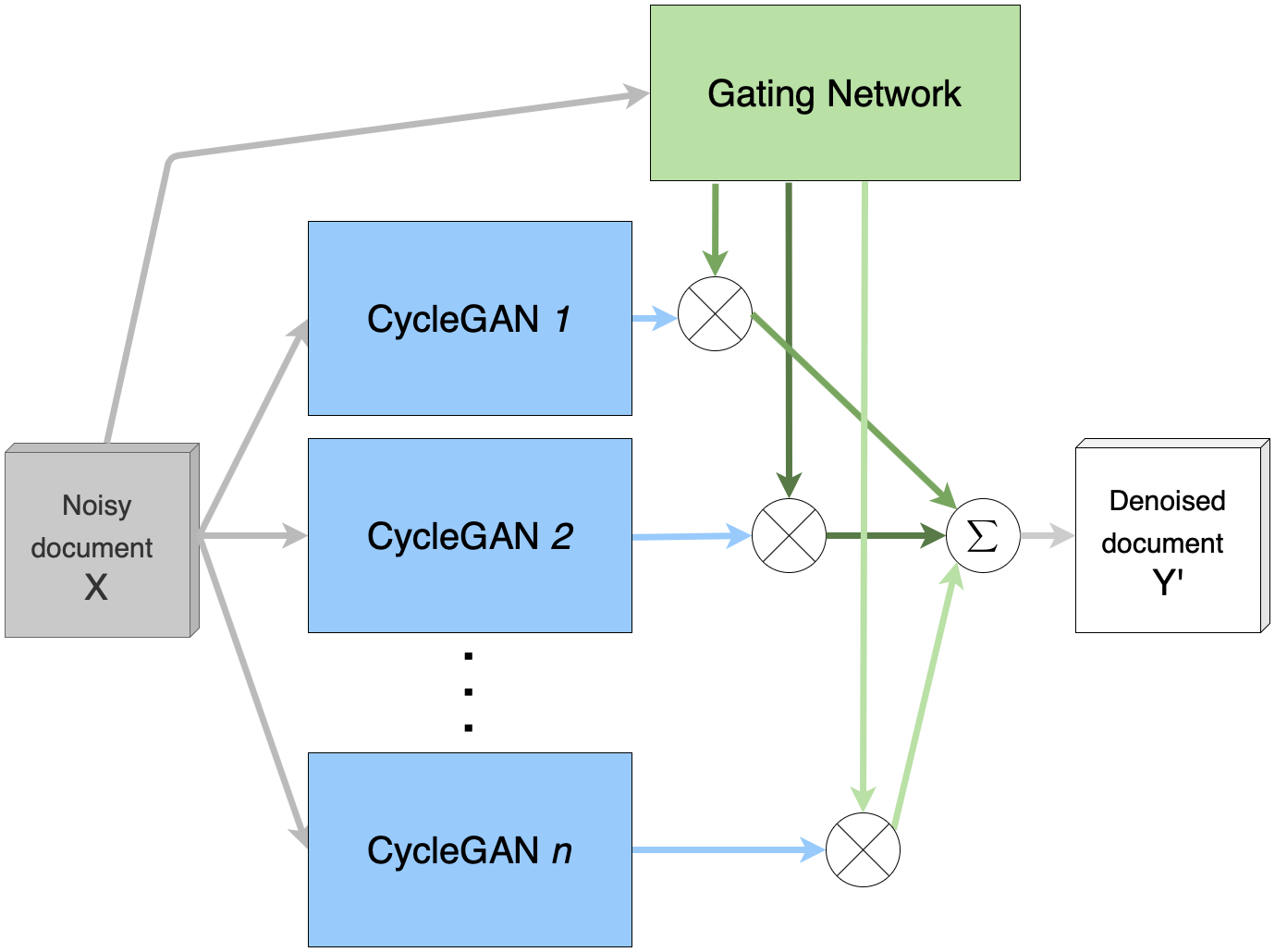}}
    \subfloat[]{\label{fig:DeepMoE_cycleGAN}
    \includegraphics[width=.7\textwidth]{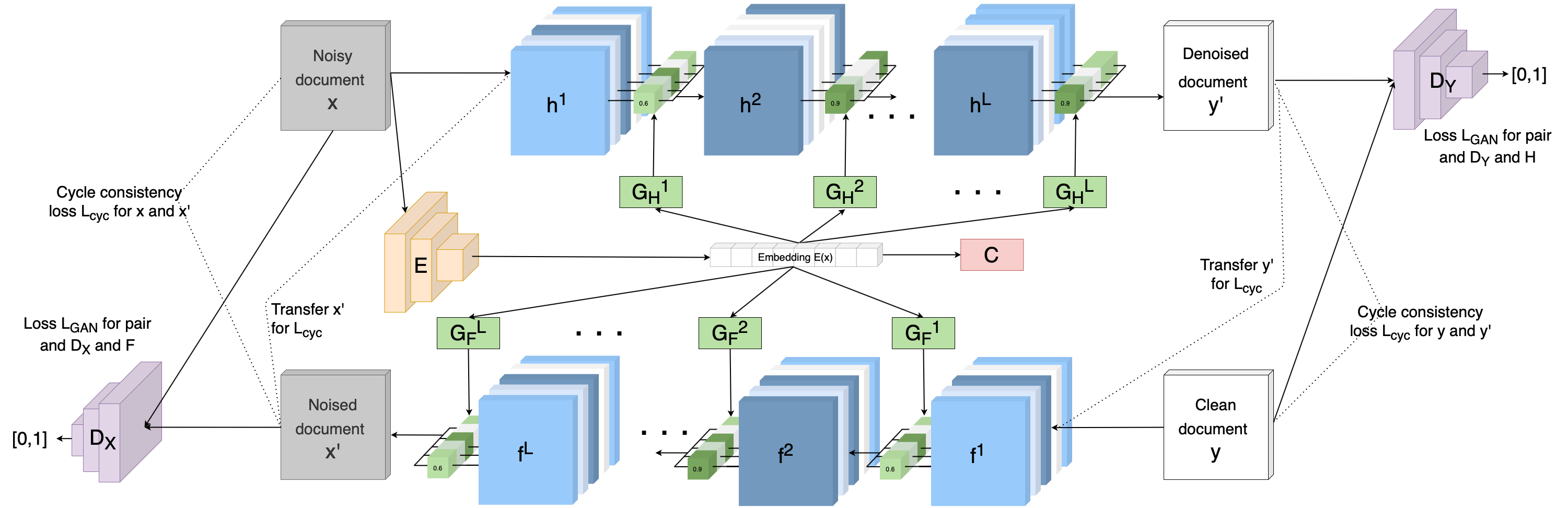}}
    \caption{(a) A na\"ive approach for mixture of experts on cycle-GANs. (b) The architecture of proposed unified model. $h^i$ and $f^i$ refer to output tensors of $i$-th convolutional layers of $H$ and $F$, respectively. \textit{Flat} rectangles are fully connected layers, while other components, \ie, $D_X$, $D_Y$, $E$, are CNNs.}
    \label{fig:architecture}
\end{figure*}

In order to design a unified model for image clean-up to remove various noise types from documents, we propose to integrate the deep MoE with cycle-GAN as the base model. Figure~\ref{fig:DeepMoE_cycleGAN} depicts the sketch of the proposed architecture.

The components of base cycle-GAN in proposed architecture include: two generators, \ie, forward $H$, and backward $F$ generators, as well as two discriminators $D_{\mathcal{Y}}$ and $D_{\mathcal{X}}$. The other components, \ie, an embedder $E$, a classifier $C$, and gating networks $g_H^*=G_H^{\{1, \dots, L_H\}}$ and $g_F^*=G_F^{\{1, \dots, L_F\}}$ ($L_H$ and $L_F$ are the number of convolutional layers in $H$ and $F$ generators, respectively) construct the elements of  the deep MoE in the architecture. 

\subsubsection{The pipeline architecture formulation}
\label{sec:example}
Our goal is to learn the generator function, which generates the cleaned/restored image $\hat{\mathbf{y}} \in \mathcal{Y}$ based on the input noisy image $\mathbf{x} \in \mathcal{X}$. Every noisy image is labeled by one type of noise, which is defined by $c_{\mathbf{x}} \in \mathcal{C}$. In our experiments, the types of noise/imperfections are S\&P noise, blurred, faded text, or watermarked document. During the pipeline training, we sample a tuple composed of a noisy image and its label $(\mathbf{x}, c_{\mathbf{x}}) \in \mathcal{X} \times \mathcal{C}$ and a clean unpaired image $\mathbf{y} \in \mathcal{Y}$\footnote{Our solution does not require target clean images during training. We use some metadata (noise type) to train the embedder.}.

The first step in the pipeline is to acquire the embedding vector $\mathbf{e}_{\mathbf{x}} \in \mathbb{R}^d$ by applying the noisy image $\mathbf{x} \in \mathcal{X}$ to the embedder, i.e. $\mathbf{e}_{\mathbf{x}} = E(\mathbf{x})$. Next, we calculate the gating network outputs $g^i_H = G_{H}^{i}(\mathbf{e}_{\mathbf{x}}) \in \mathbb{R}_{+}^{N_H^i}~for~i \in \{1, \dots, L_H\}$ and $g^i_F = G_{F}^{i}(\mathbf{e}_{\mathbf{x}}) \in \mathbb{R}_{+}^{N_F^i}~for~i \in \{1, \dots, L_F\}$, where $N_H^i$ refers to the number of output channels in the $i$-th convolutional layer in generator $H$.
The embedder network also predicts the label $\hat{c_{\mathbf{x}}}$ of the input image $\mathbf{x}$ based on the embedding vector, \ie, $\hat{c_{\mathbf{x}}} = C(\mathbf{e_x})$.

The next steps of the pipeline applies the gating networks $g^*_H$ and $g^*_F$ to the cycle-GAN generators $H$ and $F$, respectively. This generates the cleaned image $\mathbf{\hat{y}} = H(\mathbf{x}, g_H^1, \dots, g_{H}^{L_{H}})$ and the noisy image $\mathbf{\hat{x}} = F(\mathbf{y}, g_F^1, \dots, g_{F}^{L_{F}})$. 
In both generators, we replace a standard convolutional layer by an MoE convolutional layer, \ie, we calculate $h^{t+1}_o = \sum_{i=1}^{N_H^t} (g_H^t)_i \mathbf{K}^t_i * h^t_i$, where $h^t$ is an output tensor of the $t$-th convolution layer and $h^t_i$ refers to a tensor on its $i$-th channel; $*$ is a convolutional operator and $\mathbf{K}^t_i$ denotes a kernel for the $t$-th layer and the $i$-th channel.

The discriminators work in the same way as in the original cycle-GAN, i.e. $D_\mathcal{X}(\mathbf{x}) \in [0,1]$ and $D_\mathcal{Y}(\mathbf{y}) \in [0,1]$.

\subsubsection{The loss function formulation}

First, we define two additional functions to formulate cycle-GAN generators with deep MoE layers:
\begin{equation}\label{eq:FGen}
    H^{\textup{MoE}}(\mathbf{x}, \mathbf{x}) = H(\mathbf{x}, G^{1}_{H}(E(\mathbf{x})), \dots, G^{L_H}_{H}(E(\mathbf{x}))),
\end{equation}
and
\begin{equation}\label{eq:BGen}
    F^{\textup{MoE}}(\mathbf{y}, \mathbf{x}) = F(\mathbf{y}, G^{1}_{F}(E(\mathbf{x})), \dots, G^{L_F}_{F}(E(\mathbf{x}))).
\end{equation}
Note that only the noisy image $\mathbf{x}$ is provided to the embedder and gating networks. The same argument appears twice in Eq.~(\ref{eq:FGen}) in order to keep the same function structure as in Eq.~(\ref{eq:BGen}). 
Next, we formulate a novel loss function $\mathcal{L}_{\textup{GAN}}$ for $H^{\textup{MoE}}$, $D_\mathcal{Y}$, $\mathcal{X}$, $\mathcal{Y}$ 
as follows:
\begin{multline}\label{eq:GAN_Loss1}
  \mathcal{L}_{\textup{GAN}}(H^{\textup{MoE}}, D_{\mathcal{Y}}, \mathcal{X}, \mathcal{Y}) =  \mathbb{E}_{\mathbf{y} \sim \mathcal{Y}}[\log D_{\mathcal{Y}}(\mathbf{y})] \\
  + \mathbb{E}_{\mathbf{x} \sim \mathcal{X}}[1 - \log D_{\mathcal{Y}}(H^{\textup{MoE}}(\mathbf{x}, \mathbf{x}))],  
\end{multline}
Similarly, we can define the $\mathcal{L}_{\textup{GAN}}$ for $F^{\textup{MoE}}$ as follows:
\begin{multline}\label{eq:GAN_Loss2}
  \mathcal{L}_{\textup{GAN}}(F^{\textup{MoE}}, D_{\mathcal{X}}, \mathcal{Y}, \mathcal{X}) = \mathbb{E}_{\mathbf{x} \sim \mathcal{X}}[\log D_{\mathcal{X}}(\mathbf{x})] \\
  +\: \mathbb{E}_{\mathbf{y} \sim \mathcal{Y}, \mathbf{x} \sim \mathcal{X}}[1 - \log D_{\mathcal{X}}(F^{\textup{MoE}}(\mathbf{y}, \mathbf{x}))].  
\end{multline}

On the other hand, the cycle-consistency loss $\mathcal{L}_{\textup{cyc}}$ for $H^{\textup{MoE}}$ and $F^{\textup{MoE}}$ is defined as provided below:
\begin{multline}\label{eq:cycle_Loss}
\begin{split}
  & \mathcal{L}_{\textup{cyc}}(H^{\textup{MoE}}, F^{\textup{MoE}}, \mathcal{X}, \mathcal{Y}) = \\
  & \mathbb{E}_{\mathbf{x} \sim \mathcal{X}}[ \parallel F^{\textup{MoE}}(H^{\textup{MoE}}(\mathbf{x}, \mathbf{x}), \mathbf{x}) - \mathbf{x} \parallel_{1}]  \\ 
  + & \mathbb{E}_{\mathbf{y} \sim \mathcal{Y}, \mathbf{x} \sim \mathcal{X}}[ \parallel H^{\textup{MoE}}(F^{\textup{MoE}}(\mathbf{y}, \mathbf{x}), \mathbf{x}) - \mathbf{y} \parallel_{1}]. 
\end{split}
\end{multline}

The final objective function for cycle-GAN network is formulated based on a combination of losses given in Eqs.~(\ref{eq:GAN_Loss1}), (\ref{eq:GAN_Loss2}), and~(\ref{eq:cycle_Loss}): 
\begin{multline}\label{eq:cycleGAN_Loss}
\begin{split}
& \mathcal{L}_{\textup{cycle-GAN}}(H^{\textup{MoE}}, F^{\textup{MoE}}, D_{\mathcal{X}}, D_{\mathcal{Y}}, \mathcal{X}, \mathcal{Y}) = \\
& \mathcal{L}_{\textup{GAN}}(H^{\textup{MoE}}, D_{\mathcal{Y}}, \mathcal{X}, \mathcal{Y})  
+  \mathcal{L}_{\textup{GAN}}(F^{\textup{MoE}}, D_{\mathcal{X}}, \mathcal{Y}, \mathcal{X}) \\ 
+ & \lambda_{\textup{cyc}} \mathcal{L}_{\textup{cyc}}(H^{\textup{MoE}}, F^{\textup{MoE}}, \mathcal{X}, \mathcal{Y}).
\end{split}
\end{multline}

In addition, we formulate the loss function required for training deep MoE Layers as follows:
\begin{multline}\label{eq:MoE_Loss}
\begin{split}
&\mathcal{L}_{\textup{MoE}}(E, G_H^{*}, G_F^{*}, C, \mathcal{X}, \mathcal{C}) = \\
&\mathbb{E}_{\mathbf{x}, c_x \sim \mathcal{X}, \mathcal{C}}[\textup{CrossEntropy}(C(E(\mathbf{x}, c_\mathbf{x})))] \\
+ &\lambda_{G_{H}} \mathbb{E}_{\mathbf{x} \sim \mathcal{X}}[\sum_{l=1}^{L_H} \parallel G_H^{l}(E(\mathbf{x})) \parallel_{1}]  \\
+ &\lambda_{G_{F}} \mathbb{E}_{\mathbf{x} \sim \mathcal{X}}[ \sum_{l=1}^{L_F} \parallel G_F^{l}(E(\mathbf{x})) \parallel_{1}].
\end{split}
\end{multline}


The total and final objective function for training the integrated cycle-GAN and deep MoE Layers is derived by combining the losses given in Eqs.~(\ref{eq:cycleGAN_Loss}) and~(\ref{eq:MoE_Loss}):
\begin{multline}
\mathcal{L}_{\textup{cycle-GAN}}(H^{\textup{MoE}}, F^{\textup{MoE}}, D_{\mathcal{X}}, D_{\mathcal{Y}}, \mathcal{X}, \mathcal{Y}) \\
+ \lambda_{\textup{MoE}} \mathcal{L}_{\textup{MoE}}(E, G_H^{*}, G_F^{*}, C, \mathcal{X}, \mathcal{C}).
\end{multline}

\section{Implementation details}

\subsection{Model architecture}
The cycle-GAN network is adopted from the implementation explained in~\cite{Zhu2017}. 
ResNets of nine blocks are used in the generators. The discriminator networks are $70 \times 70$ Patch-GANs~\cite{cGAN2017}, which classify $70 \times 70$ overlapping patches as real or fake. The embedder network is a 7-layer CNN with kernel size of $3 \times 3$, batch normalization and ReLU activation functions. The classifier $C$, which is the last layer of the embedder is a fully connected layer with softmax and cross-entropy loss function for classification of the input patches based on their noise type (4 classes). Finally, the gating networks are fully connected networks with ReLU activation functions, whose inputs are from the penultimate layer of the embedder, \ie, just before the classifier $C$. There are 18 gating networks for the two generators, 9 for each generator corresponding to the number of blocks per generator. 
It is worthwhile to highlight here that these networks are essential during training of the model. For the inference, \ie, for document image clean-up, a minimal model is required, including the forward generator $H^{\textup{MoE}}$, the embedder $E$, and the gating networks $g^*_H$ corresponding to the forward generator $H^{\textup{MoE}}$. This greatly reduces the network complexity during the inference in terms of network latency and memory footprint.

\subsection{Hyper-parameters and model training}

Our experiments showed that the same as the original cycle-GAN~\cite{Zhu2017}, using a least-square loss instead of the negative log likelihood or Wasserstein loss for $\mathcal{L}_{GAN}$ results in a more stable training and a better performance in document image clean-up application, and hence we adopted this loss during training the model. Also, same as the original cycle-GAN, as suggested by Shrivastava \etal~\cite{Shrivastava2017}, the discriminators were updated based on a history of 50 previously generated images instead of just the last generated image by the generators. We empirically set $\lambda_{\textup{cyc}}=10$, $\lambda_{\textup{MoE}}=1$, and $\lambda_{G_{F}}=\lambda_{G_{H}}=0.1$. Adam optimizer with the batch size of 32 was used with a learning rate of $2\times10^{-4}$. We used 4 Nvidia Tesla V100 GPUs for training the model, which took about 1.5 days to complete. The model was implemented using PyTorch framework.

\begin{table*}[!t]
    \caption{The details of the dataset used for training the proposed model.}
    \centering
    \label{tab:training_dataset}
    \begin{tabular}{|p{4cm}|c|c|c|c|c|}
        \hline
        Document Types    & \multicolumn{3}{c|}{Lease Contracts} & Tax Forms & Invoices\\ \hline \hline
        Noise Types       & S\&P & Blurred    & Faded & Watermarked   & Watermarked \\ \hline
        No. of noisy/clean pages  & 663/1125   & 1513/1125  & 377/1125  & 5416/5416 & 1339/1339   \\ \hline
        Page Size / Patch Size  & \multicolumn{5}{c|}{Resized to the Closest Multiple of Patch Size / $256 \times 256$} \\ \hline
        Total No. of Noisy Patches & 290,119 & 578,316  & 137,181   &  \multirow{2}{*}{2,212,729}  & \multirow{2}{*}{542,121}\\ \cline{1-4}
        Total No. of Clean Patches & \multicolumn{3}{c|}{440,159} & ~  & ~   \\ \hline
        Data Augmentation   & \multicolumn{5}{c|}{Overlapping with the Stride of 128 Pixels} \\ \hline
    \end{tabular}
\end{table*}  
    
\section{Experimental setup and results}
\subsection{Training dataset}

There are three main document types: unstructured (such as lease contracts, and scientific papers), semi structured (\eg, invoices), and structured (like tax forms). In order to prepare a \emph{training set}, document pages of various types and noise contents were selected from our in-house documents. The most common noise types on lease contracts are S\&P noise, blurred, or faded text, whereas tax forms and invoices are mostly in digital format containing watermarks. The number of pages in each category is shown in Table~\ref{tab:training_dataset} along with other details about the dataset. Overlapping patches of $256 \times 256$ pixels with the stride of 128 pixels were extracted from these pages to train the network. 

The set of noisy and clean pages for the lease contracts are completely unpaired. As for the tax forms and invoices, extracting patches of size $256 \times 256$ pixels from the original watermarked pages results in only 10\% patches with watermark (since only a small part of the page is watermarked). Therefore, we synthetically added watermarks to the grids of $4 \times 2$ of clean tax forms and invoices with the variations in fonts, text, size, orientation (0, $\pm$45), colors (gray, light gray, red, purple, and blue), and transparency (scale of 0.1 to 0.6), randomly selected from a uniform distribution (a sample page is shown in Figure~\ref{fig:watermarked_original}).
Although this approach generates pairs of watermarked/clean pages, we have not used this information in training the proposed model as the patches were randomly selected from the sets of watermarked and clean pages for training the model irrespective of how they were paired. 

\subsection{Test datasets} In order to assess the performance of trained unified model, three datasets are used.

\noindent {\bf Dataset I}: 100 clean and high quality pages from scientific papers. This provides more tightly controlled conditions for quantitative assessment as the OCR on the original high-quality pages can be considered as the ground truth. Watermarks were synthetically added to each page.

\noindent {\bf Dataset II}: 100 pages selected from Tobacco800 dataset~\cite{SignatureDetection-CVPR07}. These pages were originally noisy and were cleaned using our unified image clean-up approach. 

\noindent {\bf Dataset III}: A dataset of 300 in-house noisy documents containing various noise types, including S\&P noise, blurred/faded text, and watermarks.




\subsection{Evaluation metrics}
Since the ultimate purpose of image cleansing is improving the performance of the OCR, 
we used the improvement in OCR as the metric for the quantitative assessment of the model. We used ABBYY FineReader12 as the OCR engine, and OCRed both the original noisy document pages as well as cleaned ones (outputs of the model). Since the ground truth for the characters are not available, we extracted words from characters on each page and used a \emph{relative metric}: the words found on the cleaned page were considered as the reference and compared with the words on the noisy page. The percentage of mismatches between the two was then computed as a metric to measure the amount of improvement. In case that the original clean pages are available (\eg, in Dataset I), we compared them as a reference with the noisy (watermarked in Dataset I) pages as well as cleaned pages. We then provided the relative metric as a measure of deterioration. Although this is not a perfect metric, it provides a reasonable quantitative assessment in the absence of ground truth for characters. 
We reported the averaged percentage improvement/deterioration, maximum improvement/deterioration, and the percentage of pages improved/deteriorated more than 5\% and 10\%. 

In addition, since for Dataset I, the original clean and high quality images are available, we have provided the peak signal-to-noise ratio (PSNR) metric as well.

\subsection{Ablation study}

\begin{figure}[!tb]
    \centering
    \subfloat[]{\label{fig:gate_response_overall} 
    \includegraphics[width=.23\textwidth]{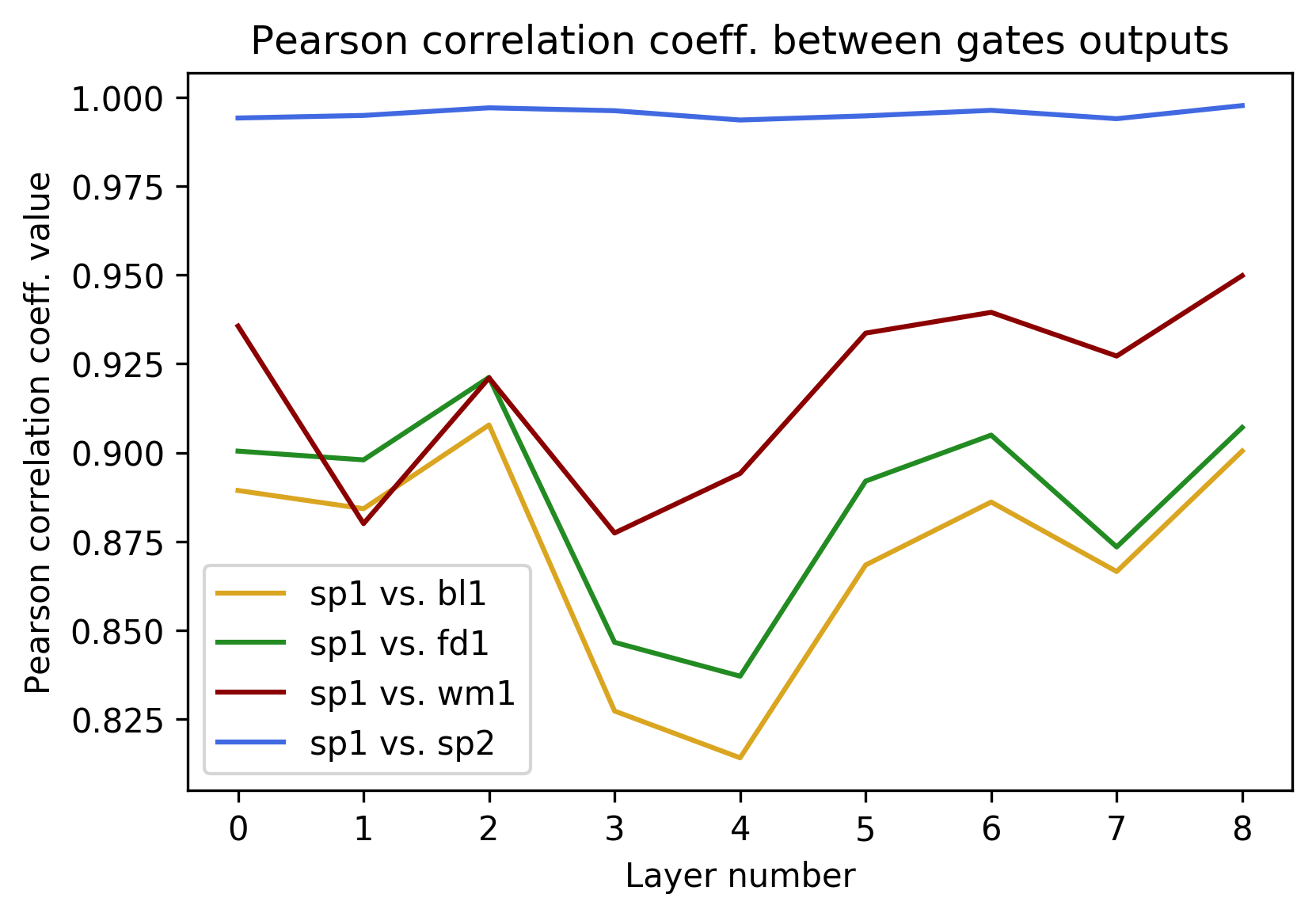}}
    \subfloat[]{\label{fig:gate_response_layer3}
    \includegraphics[width=.23\textwidth]{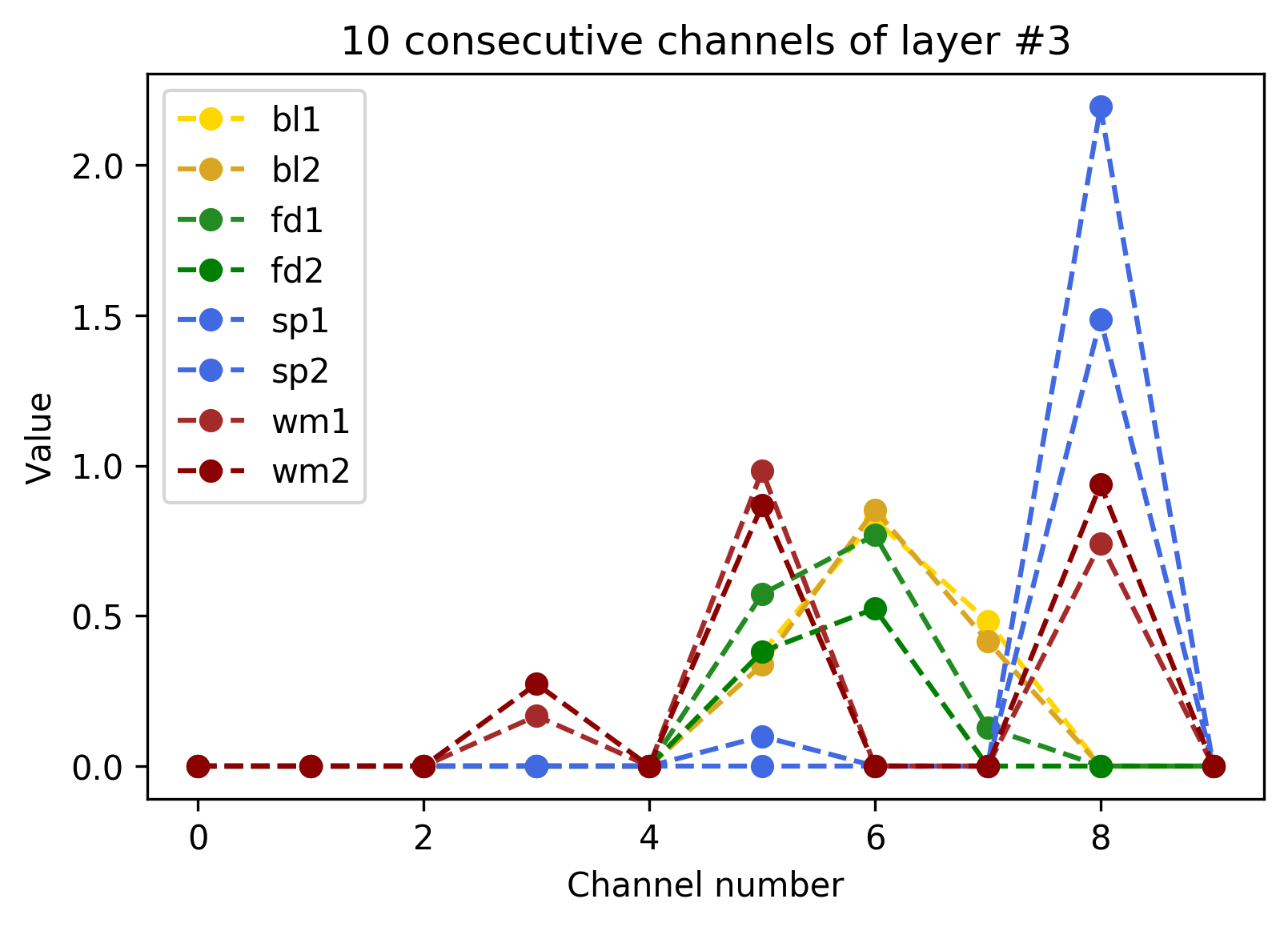}}
    \caption{The effectiveness of gating networks on various noise types used in this paper. See text for explanations.}
    \label{fig:ablation}
\end{figure}

In order to demonstrate the effectiveness of gating networks, we have provided the visualization of their outputs. In Figure~\ref{fig:gate_response_overall}, the Pearson correlation coefficients are calculated between vector outputs of forward gating network $g^*_H$ for an input image containing S\&P noise (sp1), and another image containing any other noise type (S\&P, blurred, faded, or watermarked). The correlations between two samples containing S\&P are close to one for all layers, whereas these correlations are much lower between S\&P and other noise types. Figure~\ref{fig:gate_response_layer3} displays 10 consecutive values of a section of gating network $g^*_H$ for the third convolutional layer of forward generator\footnote{The responses for all layers are provided in the Supplementary.}. These values are displayed for two samples of every considered noise type. Both these two figures demonstrate that the gating network has similar responses for the same input noise types and different patterns for different noise types. Note that in Figure~\ref{fig:gate_response_layer3} some values are 0, which is the result of $\ell_1$ loss on the gating networks. 


\subsection{Qualitative results}
Here, we present sample outputs of the trained model on unseen test patches and pages for various noise and document types\footnote{More results are provided in the Supplementary.}. Figure~\ref{fig:cleanedPatch} depicts samples of cleaned patches (along with the corresponding inputs of noisy patches) for four different artifacts, including S\&P noise, blurred, faded text, and watermark. As can be observed from these results, the proposed trained model can efficiently clean patches containing various artifacts at various intensity levels.  

Figure~\ref{fig:cleanedPage} displays samples of cleaned pages along with the corresponding inputs of noisy pages.
The rightmost images are cleaned pages generated by the proposed approach. The images in the middle are the cleaned pages generated by models trained on individual noise types (one cycle-GAN per noise type), except for the watermarked page where the middle image is cleaned using REDNet~\cite{RedNet2016,REDNET2019}. The proposed model is able to remove all noise types without distorting the contents of the page as effectively as individual models solely trained for one noise type. Also, for the watermarked page, the proposed approach was able to remove watermark as good as REDNet, which is a supervised approach and solely trained for watermark removal. 

\begin{figure*}[!tbh]
    \centering
    \subfloat[]{\label{fig:s&p_patch} \includegraphics[width=.37\textwidth]{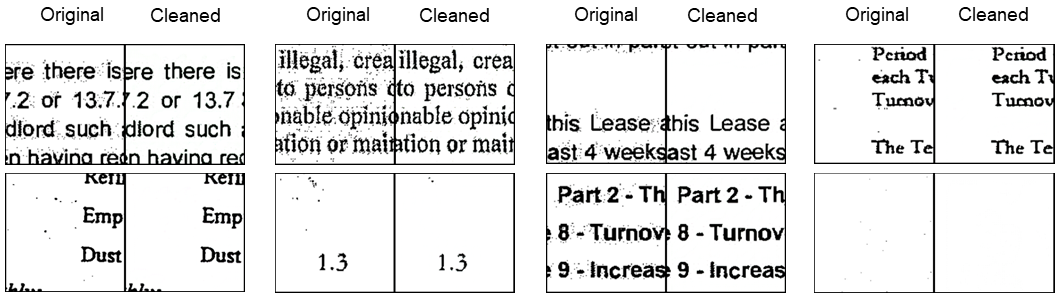}} 
    \subfloat[]{\label{fig:blurred_patch}
    \includegraphics[width=.37\textwidth]{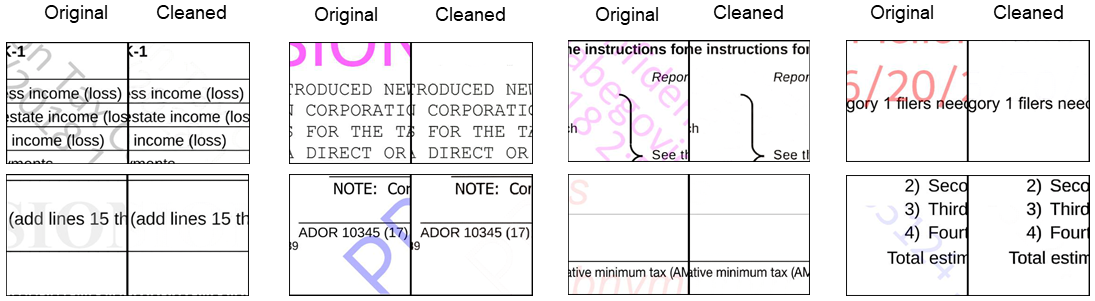}} \\ \vspace*{-0.9em}
    \subfloat[]{\label{fig:watermark_patch}
    \includegraphics[width=.37\textwidth]{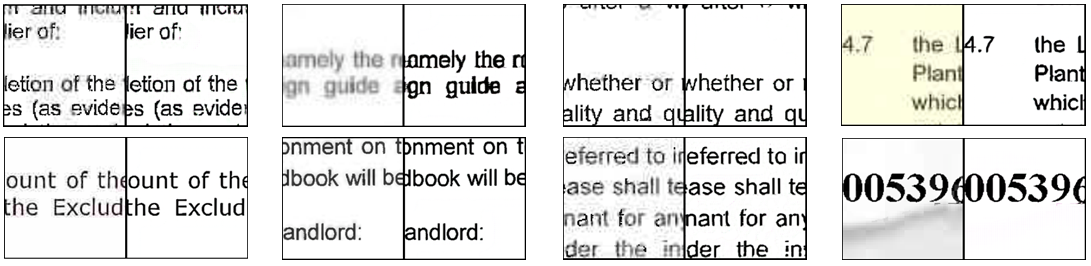}}     
    \subfloat[]{\label{fig:faded_patch} \includegraphics[width=.37\textwidth]{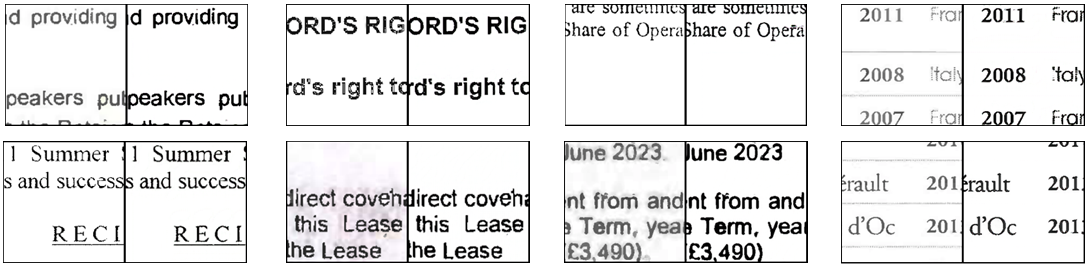}}
    \caption{The noisy inputs and cleaned outputs of the trained model at patch level for patches with (a) S\&P noise, (b) watermark, (c) blurred, and (d) faded text (best seen in digital format and zoomed in).}
    \label{fig:cleanedPatch}
\end{figure*}


\begin{figure*}[!tbh]
    \centering
    \subfloat[]{\label{fig:s&p_original} \includegraphics[width=.17\textwidth]{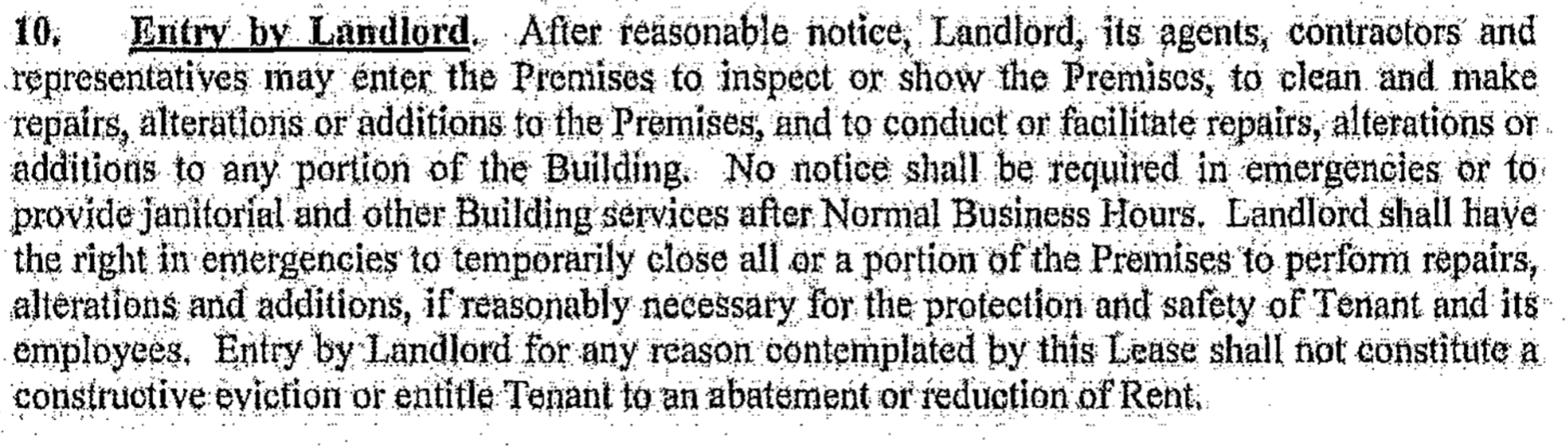}}
    \subfloat[]{\label{fig:s&p_cleaned_ind} \includegraphics[width=.17\textwidth]{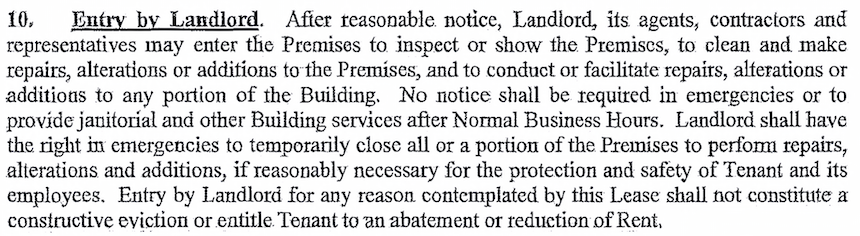}}
    \subfloat[]{\label{fig:s&p_cleaned_moe} \includegraphics[width=.17\textwidth]{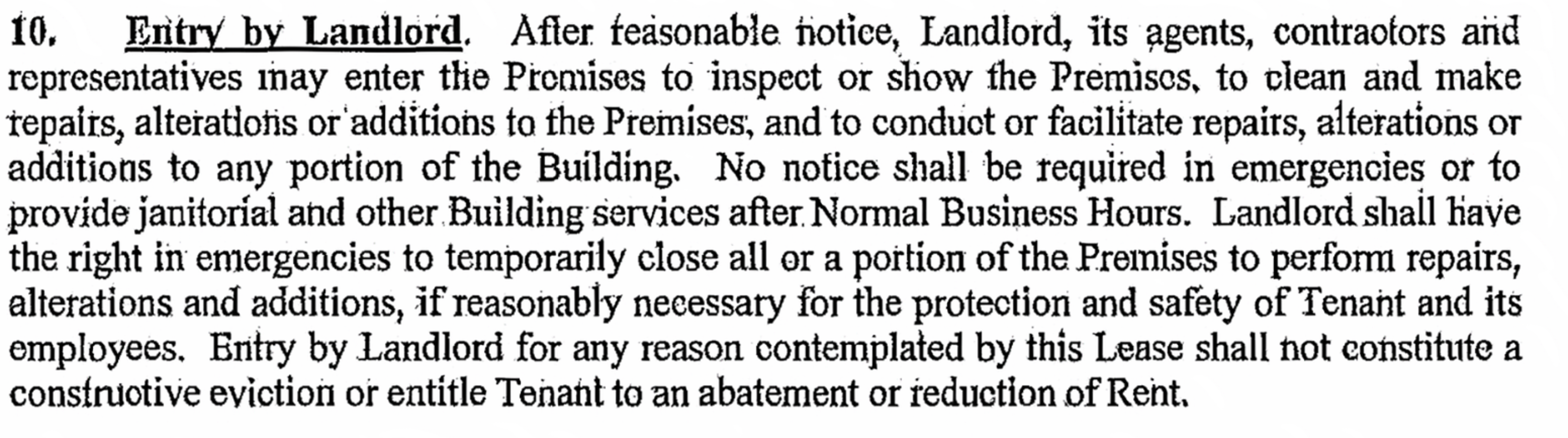}}\\ \vspace*{-0.9em}
    \subfloat[]{\label{fig:blurred_original} \includegraphics[width=.17\textwidth]{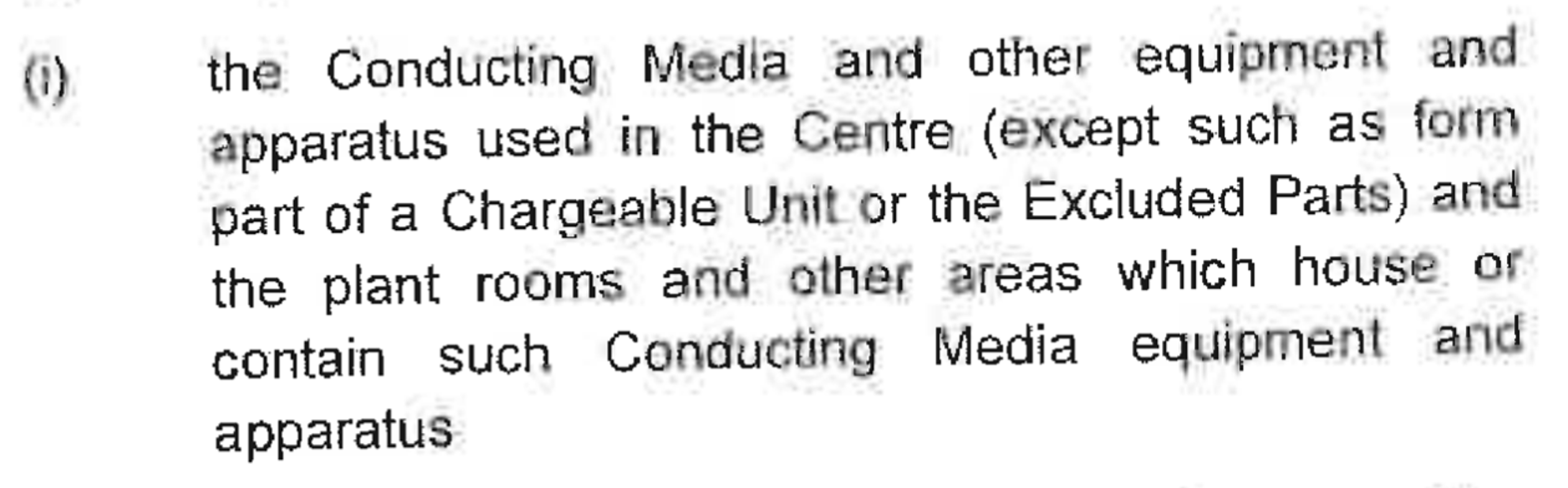}}
    \subfloat[]{\label{fig:blurred_cleaned_ind} \includegraphics[width=.17\textwidth]{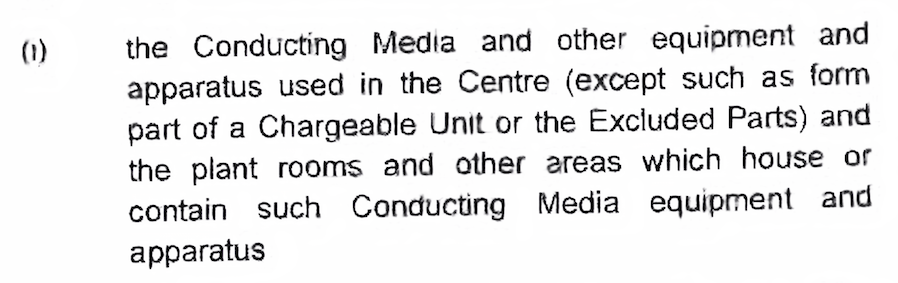}}
    \subfloat[]{\label{fig:blurred_cleaned_moe} \includegraphics[width=.17\textwidth]{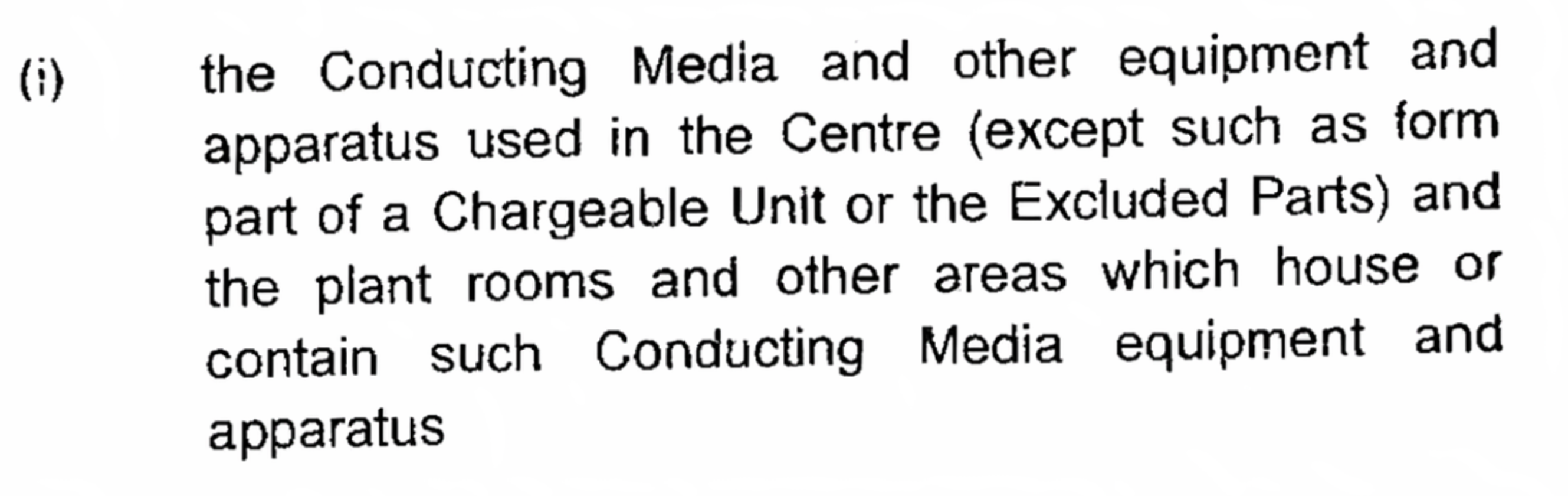}} \\ \vspace*{-0.9em}
    \subfloat[]{\label{fig:faded_original} \includegraphics[width=.17\textwidth]{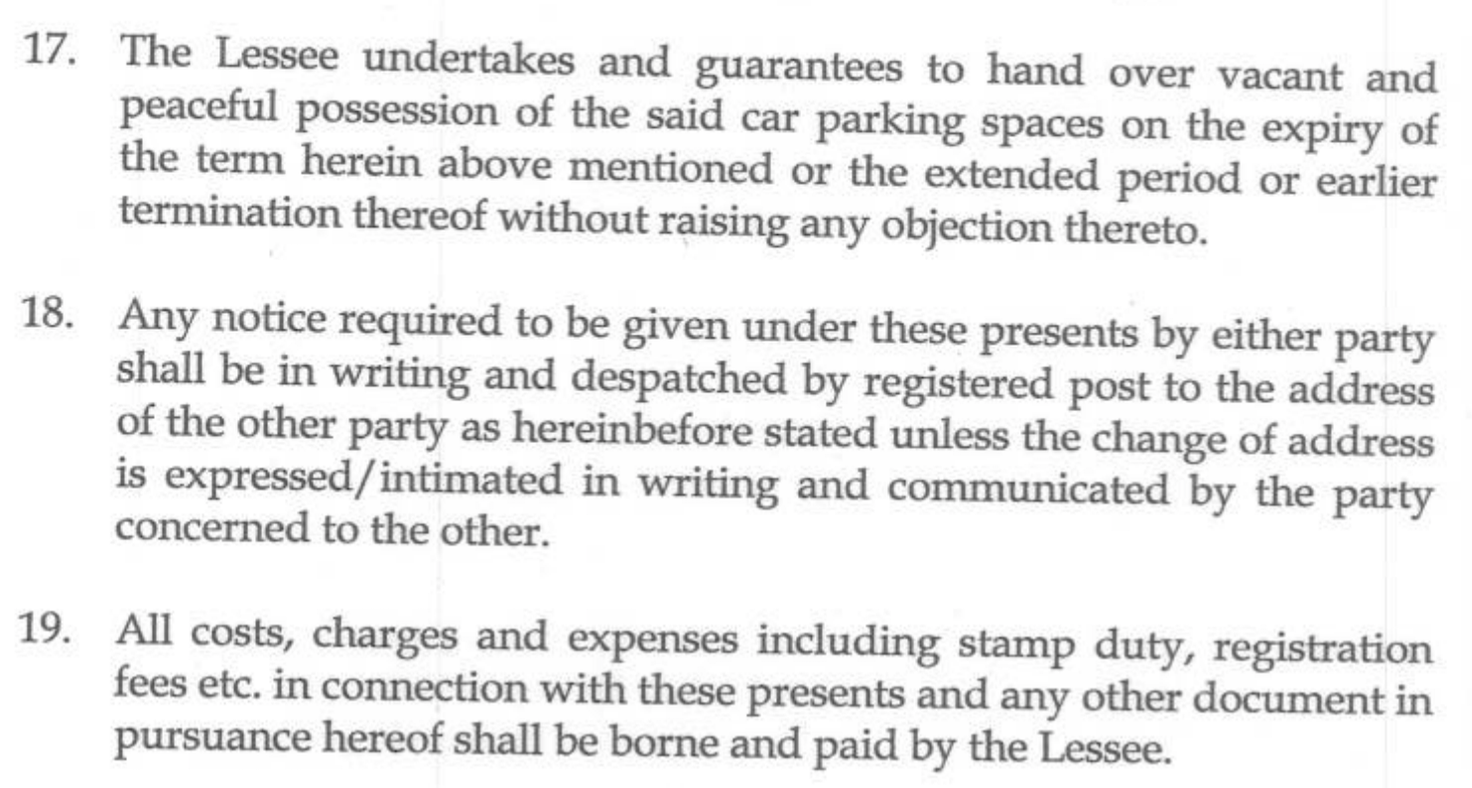}} 
    \subfloat[]{\label{fig:faded_cleaned_ind} \includegraphics[width=.17\textwidth]{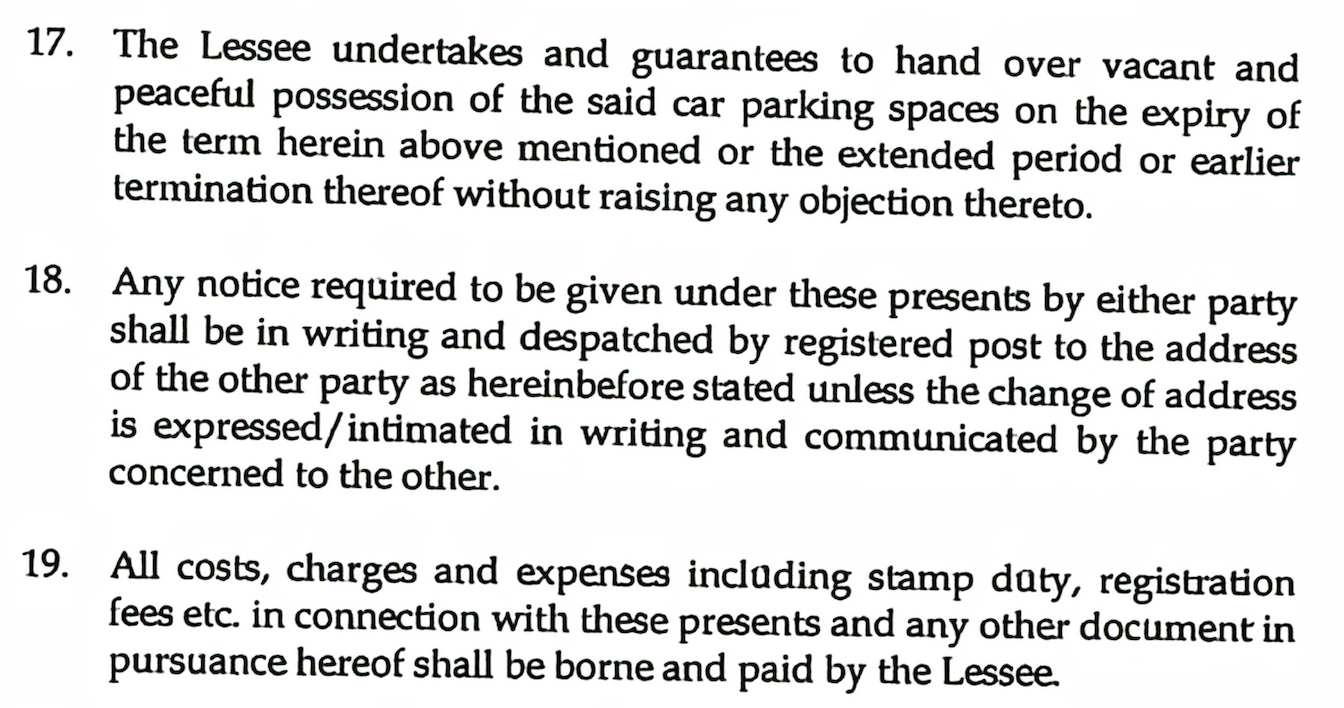}}
    \subfloat[]{\label{fig:faded_cleaned_moe} \includegraphics[width=.17\textwidth]{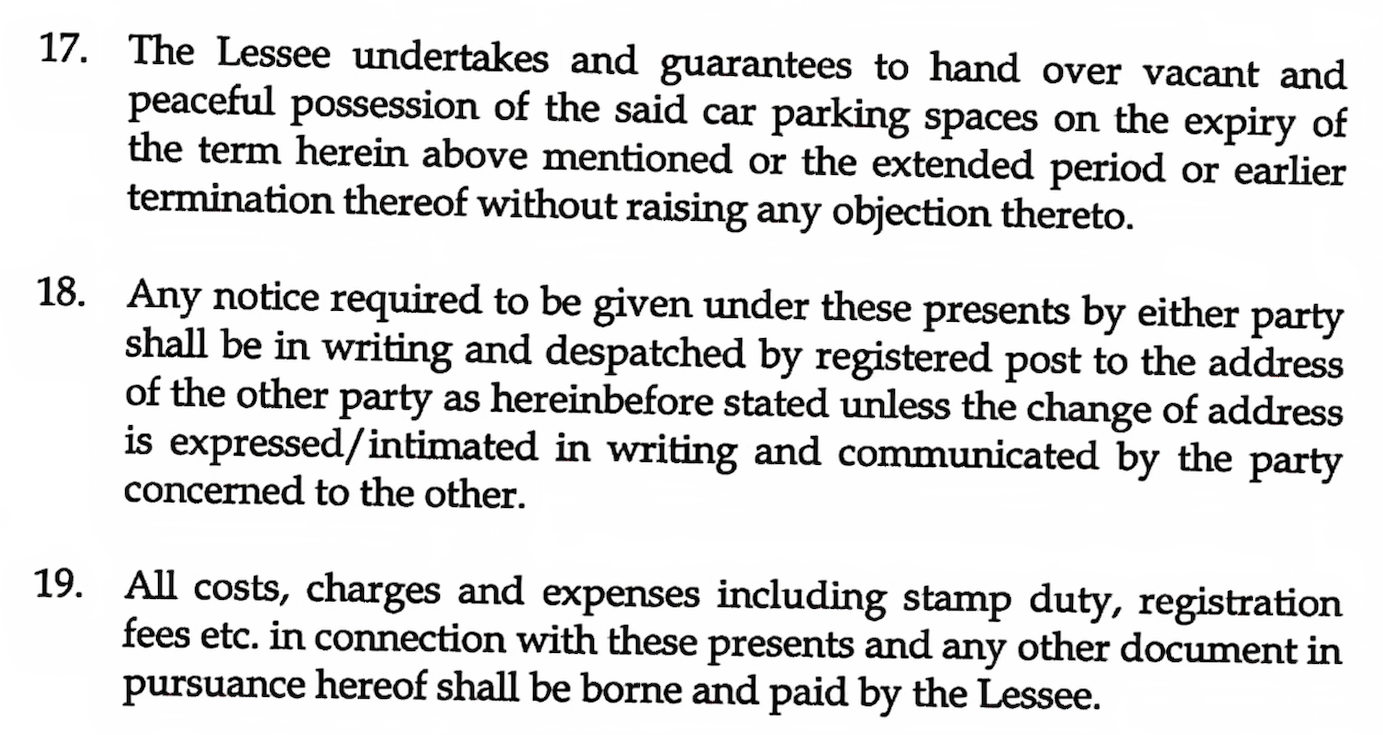}} \\ \vspace*{-0.9em}
    \subfloat[]{\label{fig:watermarked_original} \includegraphics[width=.17\textwidth]{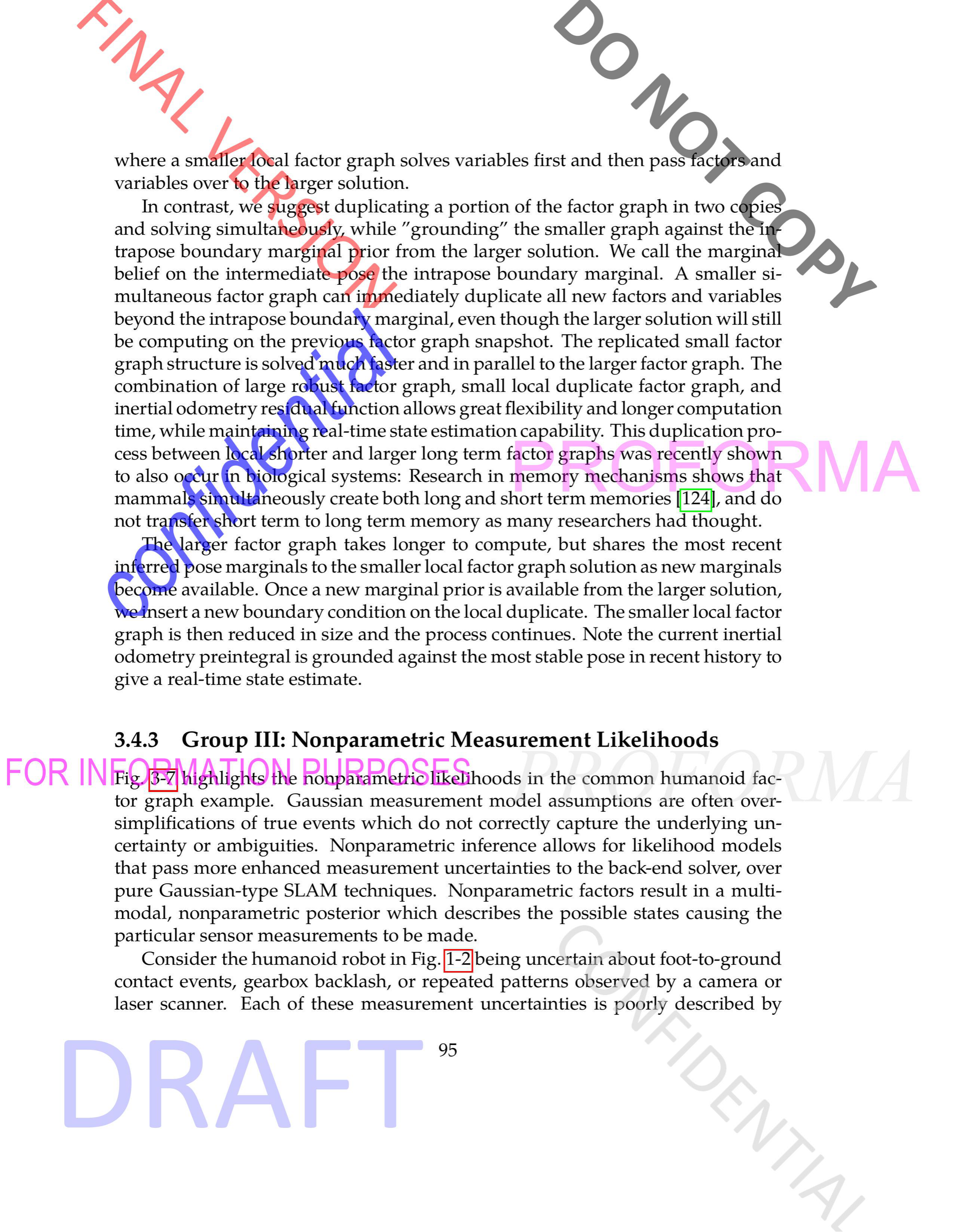}}
    \subfloat[]{\label{fig:watermarked_cleaned_REDNet} \includegraphics[width=.17\textwidth]{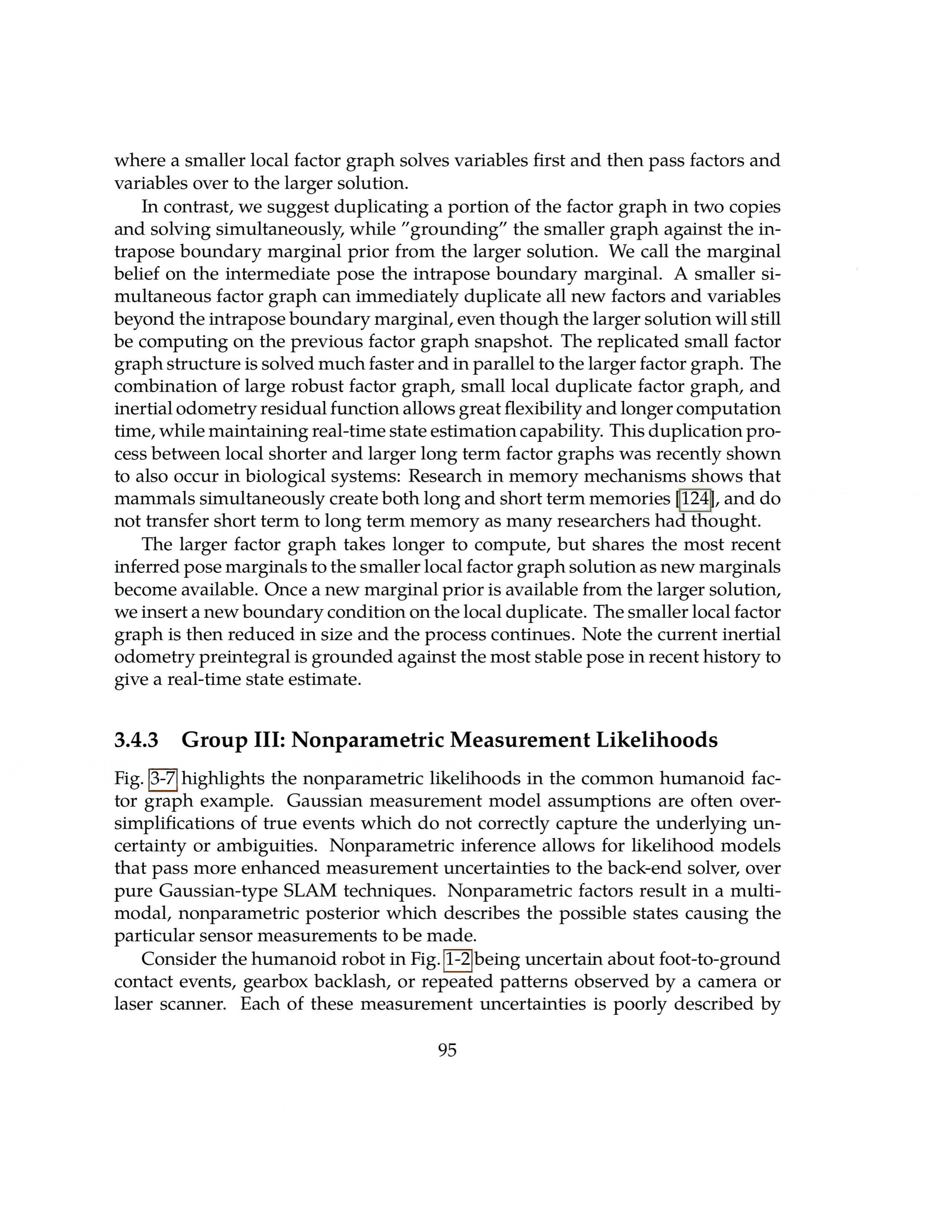}}
    \subfloat[]{\label{fig:watermarked_cleaned_moe} \includegraphics[width=.17\textwidth]{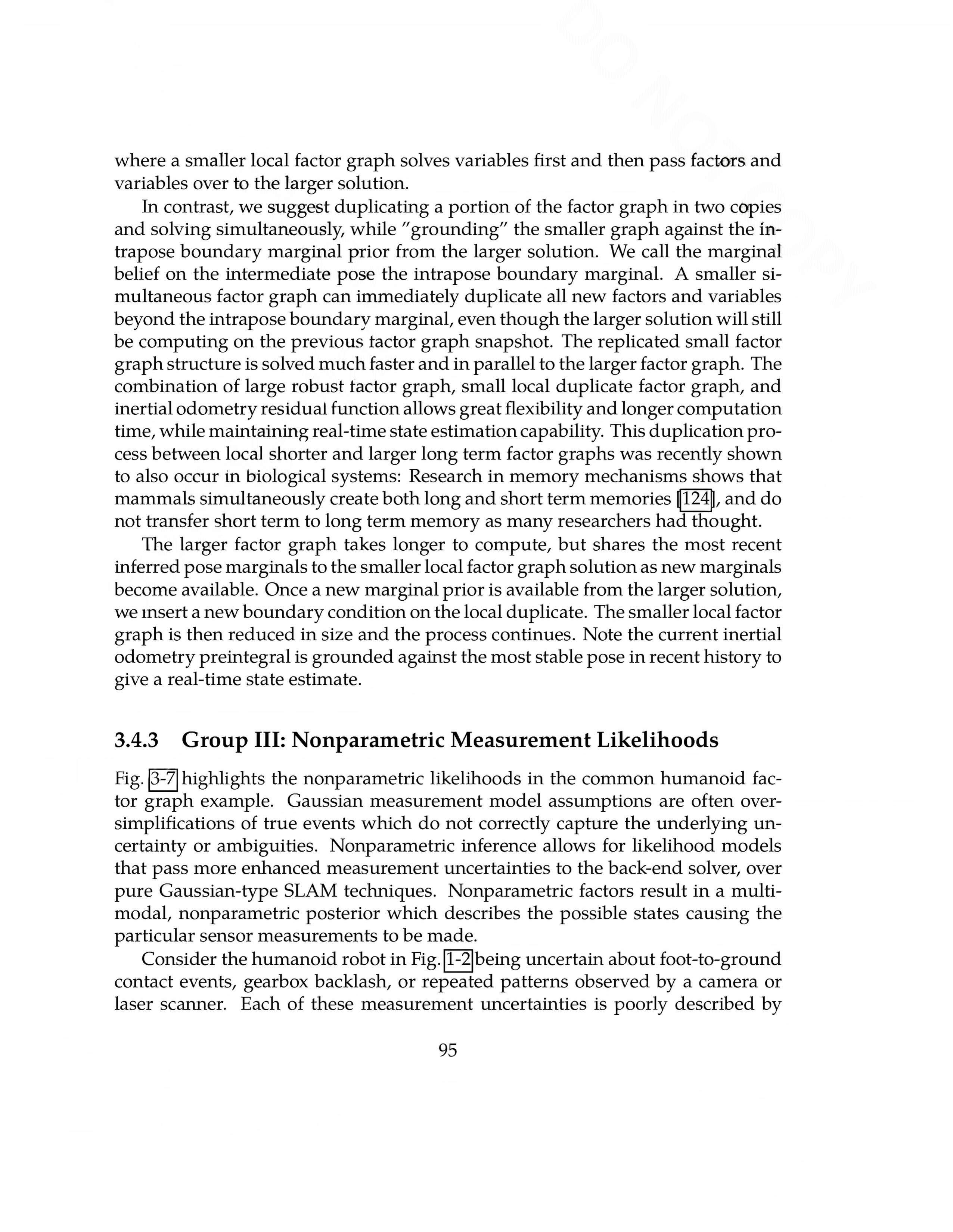}}
    \caption{The noisy inputs (left images), cleaned outputs using individual model (middle images) or REDNet (for watermarked page), as well as the proposed unified model (right images) at page level for pages with (a) S\&P noise, (d) blurred text, (g) faded text, and (j) watermark (best seen in digital format and zoomed in).}
    \label{fig:cleanedPage}
\end{figure*}

\begin{table*}[!t]
    \caption{The results of quantitative assessment of the proposed model using the \emph{relative metric} and PSNR on Dataset I. The best results, which in this case are lower numbers for OCR-based metrics and higher numbers for PSNR, as \emph{cleaned} documents are compared with the \emph{original clean} ones, are highlighted. }
    \centering
    \label{tab:results_I}
    \begin{tabular}{| L{1.2cm} |L{2.9cm}|C{1.9cm}|C{1.3cm}|C{1.3cm}|C{1.8cm}|C{2cm}|C{1.4cm}|}
        \hline
        \multicolumn{2}{|l|}{Measures}   & \multirow{2}{1.9cm}{\centering{Original vs Watermarked (\%)}} & \multicolumn{5}{c|}{Original vs Cleaned (\%)}      \\ \cline{4-8}
        \multicolumn{2}{|l|}{~}           &   ~   & REDNet~\cite{REDNET2019} & DE-GAN~\cite{DE-GAN2020} & cycle-GAN~\cite{Sharma2018} & Proposed w/o Classifier & Proposed  \\ \hline \hline
        \multicolumn{2}{|l|}{Averaged Deterioration (\%)} & 7.71 & {\bf 1.84} & 3.74 & 7.85 & 2.54 & 2.55   \\ \hline 
        \multicolumn{2}{|l|}{Max. Deterioration (\%)}    & 53.09   & {\bf 6.99}   & 19.70 & 33.33 & 14.73 & 9.93   \\ \hline 
        \multirow{2}{1.2cm}{Perc. of Pages} & +5\% Deterioration & 70     & {\bf 3} & 14 & 67 & 12 & 10       \\ \cline{2-8}
        & +10\% Deterioration & 27    & {\bf 0}     & 5 & 34 & 3 & {\bf 0}    \\ \hline
        \multicolumn{2}{|l|}{PSNR} & 35.65    & 37.64     & 37.43 & 36.52 & 37.99 & {\bf 38.33}    \\ \hline
    \end{tabular}
\end{table*}

\begin{table*}[!t]
    \caption{The results of quantitative assessment of the proposed model using the \emph{relative metric} on Datasets II and III. The best results, which in this case are higher numbers, as \emph{cleaned} documents are compared with \emph{original noisy} ones, are highlighted.}
    \centering
    \label{tab:results_II}
    \begin{tabular}{| L{1.2cm} |L{2.9cm}|C{1.5cm}|C{1.9cm}|C{1.4cm}|C{1.5cm}|C{1.9cm}|C{1.4cm}|}
        \hline
        \multicolumn{2}{|l|}{Measures}   & \multicolumn{3}{c|}{Dataset II}  & \multicolumn{3}{c|}{Dataset III}    \\ \cline{3-8}
       \multicolumn{2}{|l|}{~}           & \multicolumn{3}{c|}{Cleaned vs Original (\%)} & \multicolumn{3}{c|}{Cleaned vs Original (\%)}   \\ \cline{3-8}
        \multicolumn{2}{|l|}{~}           & cycle-GAN~\cite{Sharma2018} & cycle-GAN (Sequential) & Proposed & cycle-GAN~\cite{Sharma2018} & cycle-GAN (Sequential) &  Proposed \\ \hline \hline
        \multicolumn{2}{|l|}{Averaged Improvement (\%)}  & 5.94 & 6.03 & {\bf 7.2}    & 6.30 & 7.10 & {\bf 9.52} \\ \hline 
       \multicolumn{2}{|l|}{Max. Improvement (\%)}   & 39.02  & 52.54 & {\bf 63.06}  & 48.17 & 57.98 & {\bf 66.94} \\ \hline 
        \multirow{2}{1.2cm}{Perc. of Pages} & +5\% Improvement & 37     & 42   & {\bf 52} & 52 & 58 & {\bf 67}   \\ \cline{2-8} 
        & +10\% Improvement & 22    & 24  & {\bf 28} & 37 & 42 & {\bf 49}   \\ \hline
    \end{tabular}
\end{table*}

\subsection{Quantitative results}
For dataset I, the OCR on the original high-quality pages are considered as the ground truth, and the OCR on watermarked pages as well as cleaned pages were compared with this ground truth. The results are provided in Table~\ref{tab:results_I}. On average, the proposed unified image clean-up improves the performance of ABBYY OCR by approximately a factor of three. The low percentage of relative change in the OCR quality when comparing the original and cleaned pages demonstrates the effectiveness of the proposed algorithm in removing watermarks, and reducing the OCR errors due to watermarks on the page. We compare our proposed approach with REDNet~\cite{RedNet2016,REDNET2019} and DE-GAN~\cite{DE-GAN2020} as two representative discriminative methods. We trained the REDNet and DE-GAN separately using watermarked tax forms as specified in Table~\ref{tab:training_dataset}, and tested them on Dataset I to compare with our results. What we should consider in this comparison is that both REDNet and DE-GAN have \emph{solely} been trained on watermarked tax forms/clean pairs in a \emph{supervised} manner, whereas the proposed approach has been trained on \emph{all} noise types (including S\&P, blurred/faded text, and watermark) and all document types in an \emph{unsupervised} fashion. However, the performance of the proposed method is just slightly inferior to the REDNet and better than DE-GAN. The proposed approach outperforms DE-GAN mainly because DE-GAN, by design, removes color from inputs. Since watermarks on pages can be in color, DE-GAN has more difficulty to remove them. In addition, to demonstrate the effectiveness of the MoE in our proposed approach to train one single model to remove multiple noise types, we compare the proposed approach with a standard cycle-GAN~\cite{Zhu2017,Sharma2018} without deep MoE, and also with our proposed approach without classifier $C$ (both trained the same way as the proposed approach). The results in Table~\ref{tab:results_I} shows that the proposed approach outperforms these two networks as well. This demonstrates that removing the classifier $C$ from the final stage of the embedder makes training the embedder less efficient, and deteriorates the overall performance of the model. The classifier $C$ only supports training the embedder network $E$, and this embedding have two important functions: (a) allows the forward generator $H^{\textup{MoE}}$ to adapt to various noise types (represented by the embedding vectors) and efficiently clean them, (b) informs the backward generator $F^{\textup{MoE}}$ the type of noise to be added to its input clean image. In the standard cycle-GAN, information about noise is stored in a denoised image. In case of multiple noise types, this confuses the backward generator and the function of embedder is to avoid this confusion. 

Neither original clean pages nor the ground truth OCR are available for Datasets II and III. However, the relative OCR performance of ABBYY on the original noisy pages and the corresponding cleaned pages is used to evaluate the performance of the proposed model. The results are shown in Table~\ref{tab:results_II}. Using the proposed approach on average, the OCR is improved by 7.2\% and 9.52\% on Datasets II and III, respectively. Also, more than 50\% of pages on both datasets receive more than 5\% improvement in the OCR as a result of cleaning the pages using the proposed unified model. As for comparison with other approaches, there is no \emph{single model} in the literature that can remove all these noise types. Also, discriminative models like REDNet or DE-GAN cannot be used for comparison as they need noisy/clean pairs for training, which are not available here. Nonetheless, We compare the proposed approach with two   alternative methods using standard cycle-GAN: 1) same as previous experiment on Dataset I, training one single cycle-GAN (without deep MoE) using a training set consisting of all patches with various noise types. 
2) \emph{Standard cycle-GAN with Sequential training}: since it is difficult to train the standard cycle-GAN using all noise types in the training set, we first trained the model on more difficult to learn noise types (\eg, S\&P) for several epochs. Then we stopped training and resumed again using patches from noise types that can be removed easier like faded/ blurred text. Although this sequential training strategy improved the performance compared to ``\emph{standard}'' training, the performance is still far below the performance of individual trained models, as well as the proposed unified model. In addition, ``\emph{sequential}'' training is heuristic and needs some trial-and-error in order to find the optimal number of steps, as well as the number of epochs in each step. 
As can be observed in Table~\ref{tab:results_II}, the proposed model significantly outperforms the alternative methods in improving the OCR quality of cleaned pages.

As a final remark, the inference latency of the proposed algorithm was benchmarked on one GPU (Nvidia Tesla V100). On average, it takes 4.49 seconds to clean a page.   

\section{Conclusion}
In this paper, we proposed an end-to-end unsupervised multi-document image blind denoising that presents a unified model to remove various artifacts from all document types, without any need for target paired cleaned pages. We formulated the loss function for the proposed model and demonstrated that it can successfully remove artifacts and improve OCR on various document types. The proposed model is of low complexity and inference latency. In future work, we will replace the embedder and the classifier components by an autoencoder-based approach in order to train the embedding vector in an unsupervised manner. 

\section*{Disclaimer}

The views reflected in this article are the views of the author and do not necessarily reflect the views of the global EY organization or its member firms.


{\small
\bibliographystyle{ieee_fullname}
\bibliography{ms}
}

\pagebreak

\begin{center}
\textbf{\large Supplementary Notes: End-to-End Unsupervised Document Image Blind Denoising }
\end{center}
\setcounter{equation}{0}
\setcounter{figure}{0}
\setcounter{table}{0}
\setcounter{section}{0}
\makeatletter
\renewcommand{\theequation}{S\arabic{equation}}
\renewcommand{\thefigure}{S\arabic{figure}}
\renewcommand{\thetable}{S\arabic{table}}
\renewcommand{\thesection}{S-\arabic{section}}

\begin{table*}[!tbh]
    \caption{The architectural details for the proposed model.}
    \centering
    \label{tab:network_details}
    \begin{tabular}{|p{4cm}|p{7cm}|p{5cm}|}
        \hline
        Networks               & Details on the Components & Loss Function \\ \hline \hline
        Generators       & ResNets (9 blocks)        & \multirow{2}{*}{GAN Loss + cycle-consistency loss} \\
        Discriminators  & $70 \times 70$ Patch-GANs   & ~     \\ \hline
        Embedder (MoE)  & CNN (7 layers, kernel: $3 \times 3$, batchnorm, ReLU) & \multirow{2}{*}{Cross-entropy loss} \\
        Classifier (MoE)      & Fully connected layer with softmax    & ~ \\ 
        Gating Networks (MoE) & Fully connected ($64 \times 256$), ReLU, 18 of these networks for the two generators) & $\ell_1$ loss \\ \hline
    \end{tabular}
\end{table*}  


\begin{table*}[!tbh]
    \caption{The details of the test datasets used for quantitative assessment of the proposed model.}
    \centering
    \label{tab:test_dataset}
    \begin{tabular}{|p{4cm}|c|c|c|c|c|c|c|}
        \hline
        Datasets    & Dataset I  & Dataset II    &\multicolumn{5}{c|}{Dataset III} \\ \cline{2-8}
        ~           & Scientific Papers & Tobacco800  & \multicolumn{3}{c|}{Lease Contracts}  & Tax Forms   & Invoices \\ \hline
        Noise Types & Watermark & Various  & S\&P & Blurred    & Faded & Watermark   & Watermark \\ \hline \hline
        No. of Pages  & 100   & 100  & 60  & 100 & 60 & 40  & 40   \\ \hline
    \end{tabular}
\end{table*} 

\section{Architecture of and training the proposed model}\label{sec:training}
Table~\ref{tab:network_details} provides a summary of the networks used in the proposed model (the details are provided in the main paper). 

As explained in the main paper, we used in-house documents, including lease contracts, invoices, and tax forms to prepare the training dataset. The most common noise types on lease contracts, which are considered as unstructured documents, are S\&P noise, blurred, or faded text, whereas tax forms (structured documents) and invoices (semi-structured documents) mostly contain watermarks.

The set of noisy and clean pages for the lease contracts are completely unpaired. As for the tax forms and invoices, extracting patches of $256 \times 256$ from the original watermarked pages resulted in only 10\% patches with watermark (due to the fact that usually small part of the page is watermarked). Therefore, submitting these patches to the model did not train it adequately for watermark removal. To remedy this problem, we added watermarks similar to what are seen on actual tax forms or invoices, \ie, with the same variations in text, font, size, orientation, transparency, and color to the grids of $4 \times 2$ of clean tax forms and invoices. This increased the number of watermarked patches by a percentage of more than 60\%. A sample page with synthetically added watermark is shown in Figure~\ref{fig:watermarked-46}. The model was trained using the training dataset for 1,700,000 iterations.

\section{Test sets used for quantitative evaluation} 
Table~\ref{tab:test_dataset} provides additional information on the test sets used for the quantitative assessment of the proposed approach as reported in the main paper. 

\section{Additional results}

\subsection{Ablation study}

\begin{figure*}[!tbh]
    \centering
    \subfloat[]{\label{fig:layer0} \includegraphics[width=.2\textwidth]{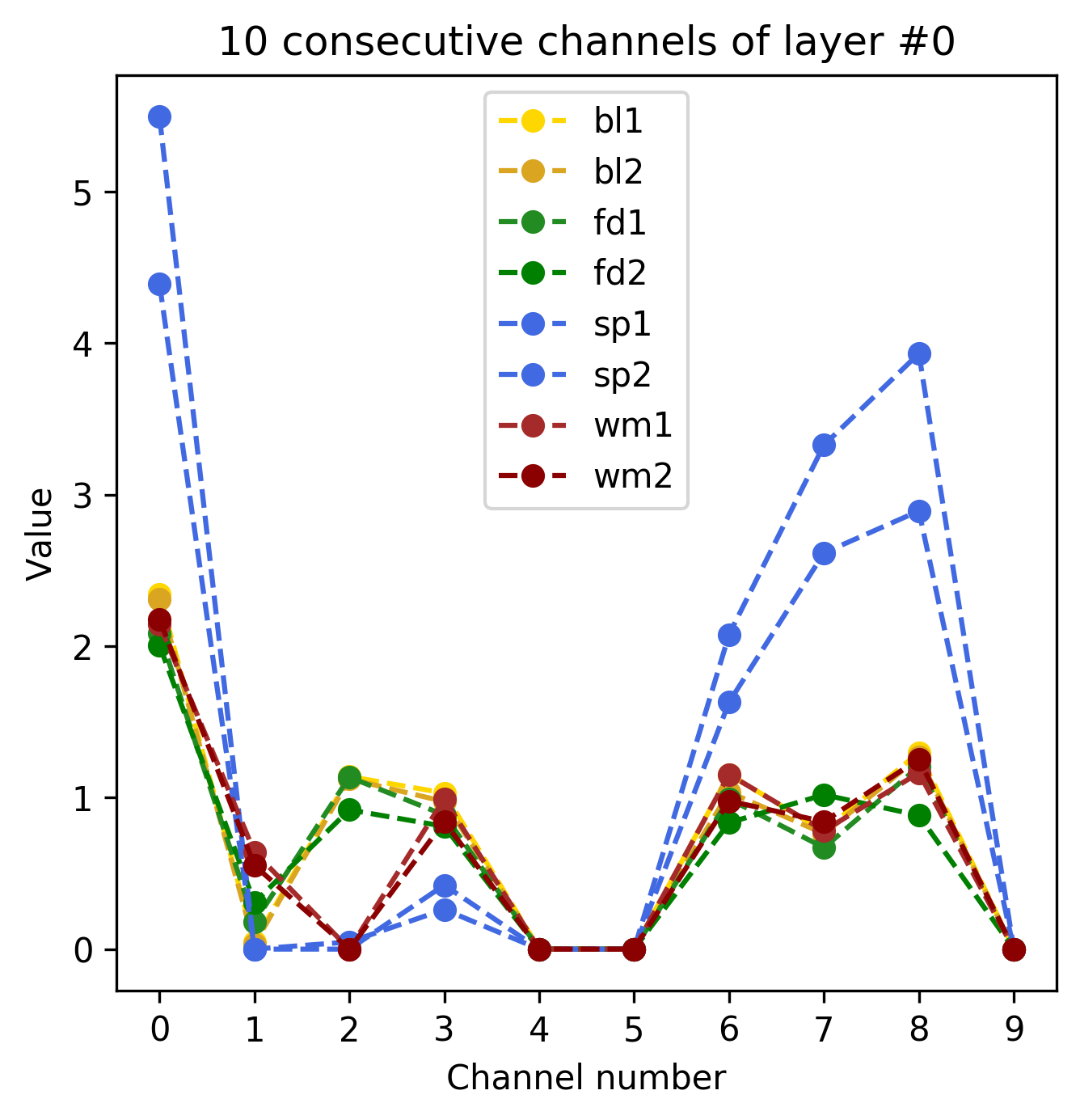}}
    \subfloat[]{\label{fig:layer1} \includegraphics[width=.2\textwidth]{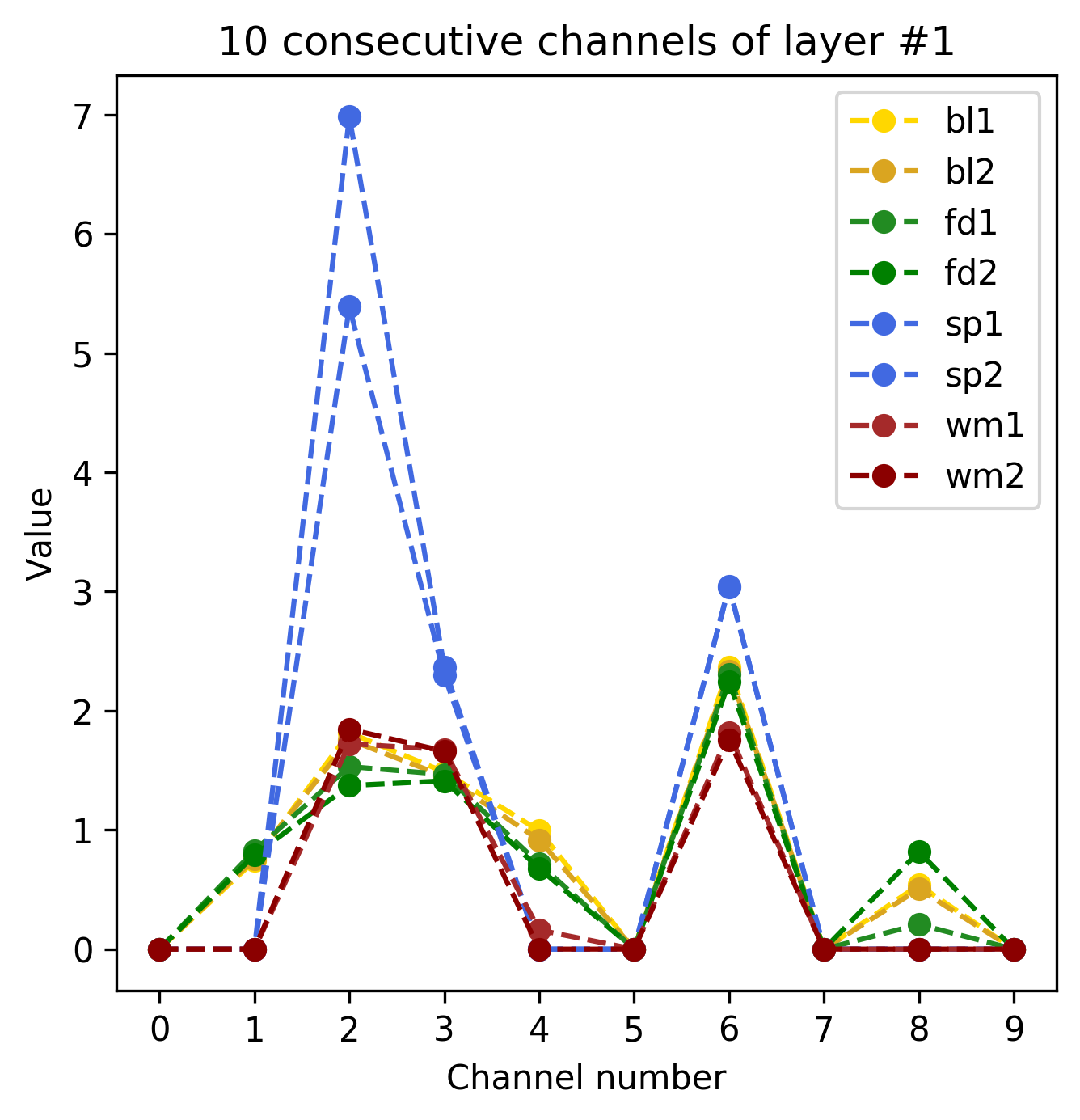}} 
    \subfloat[]{\label{fig:layer2} \includegraphics[width=.2\textwidth]{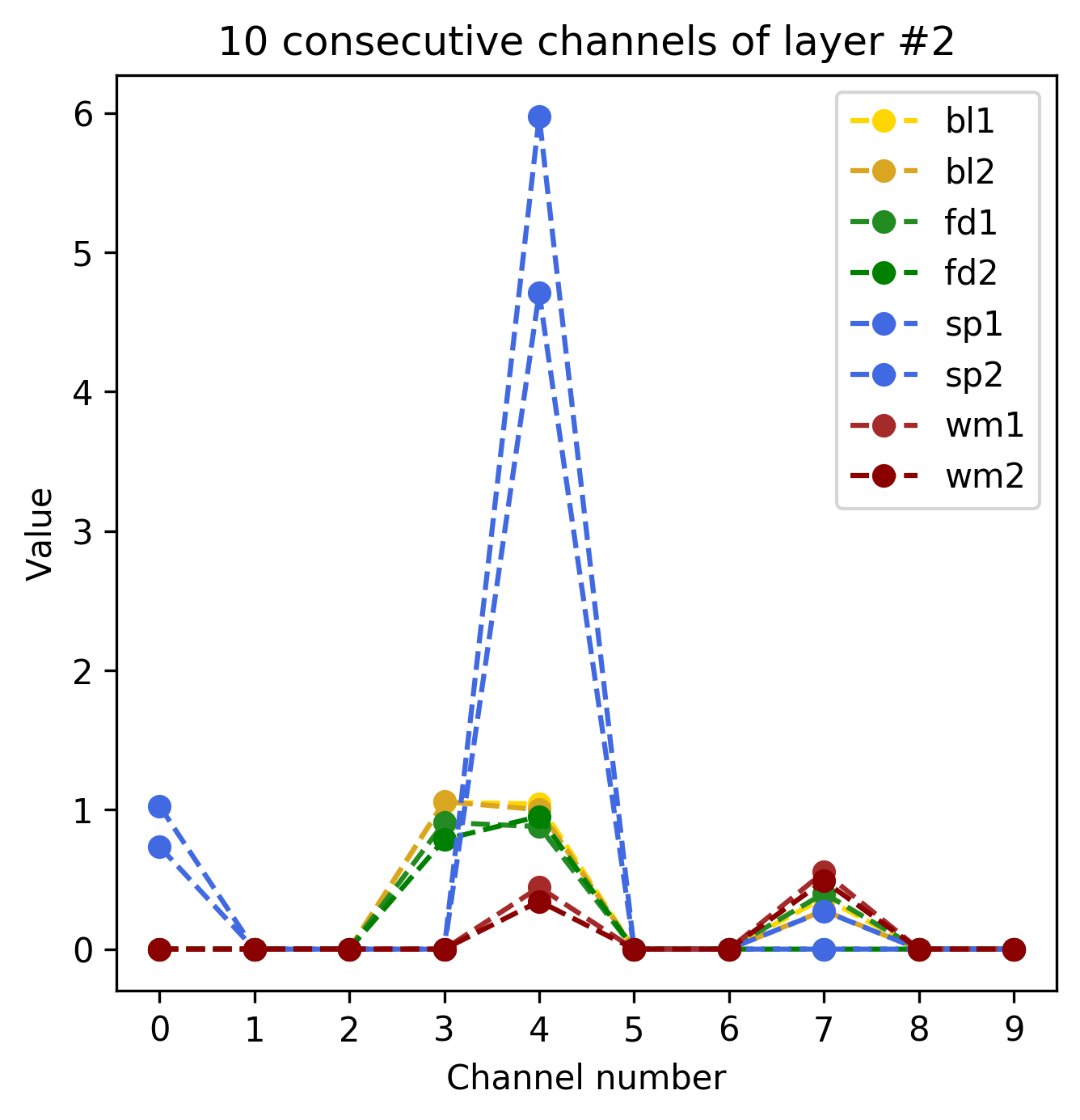}}
    \subfloat[]{\label{fig:layer4} \includegraphics[width=.2\textwidth]{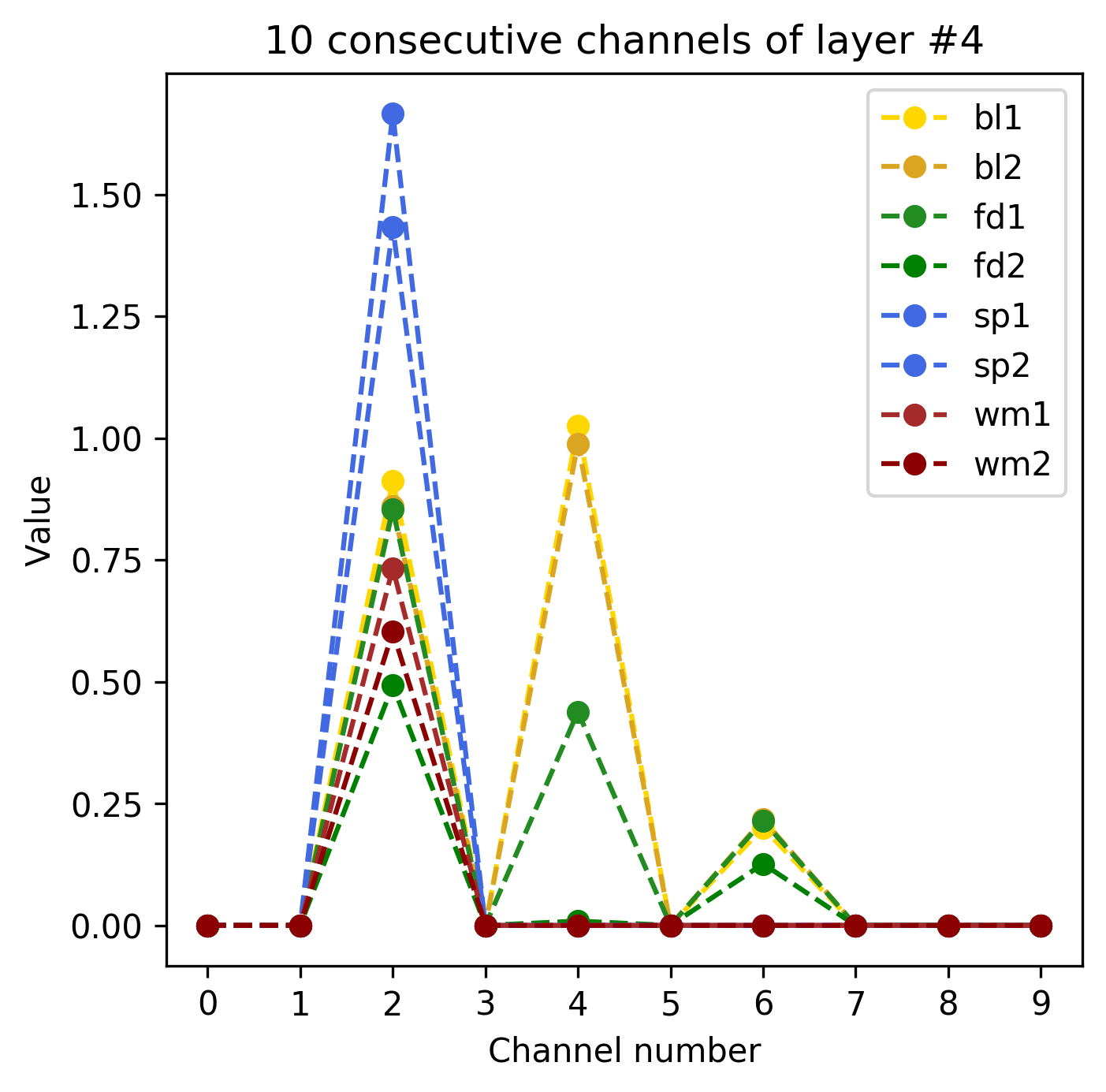}} \\ \vspace*{-0.9em}
    \subfloat[]{\label{fig:layer5} \includegraphics[width=.2\textwidth]{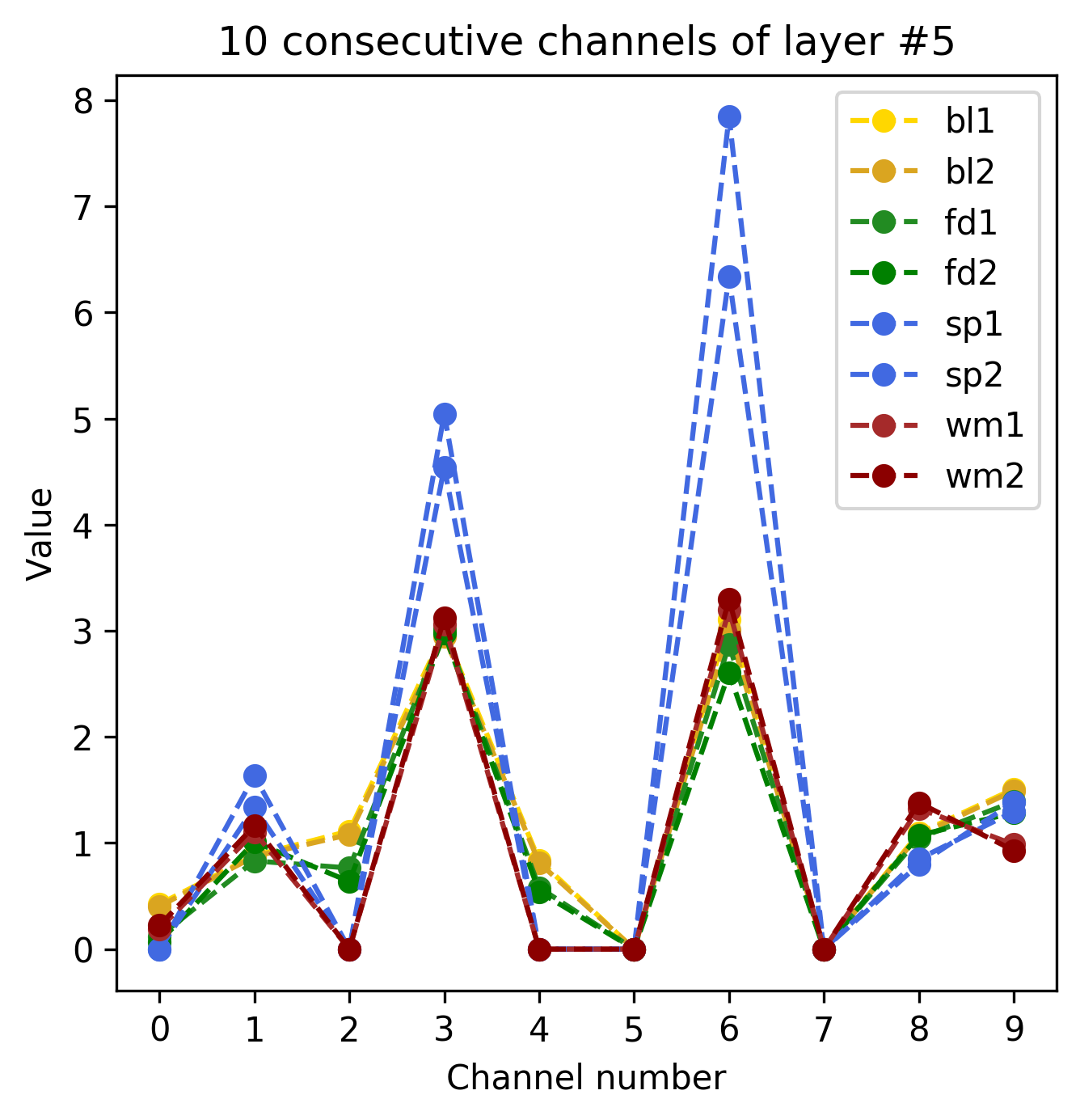}}
    \subfloat[]{\label{fig:layer6} \includegraphics[width=.2\textwidth]{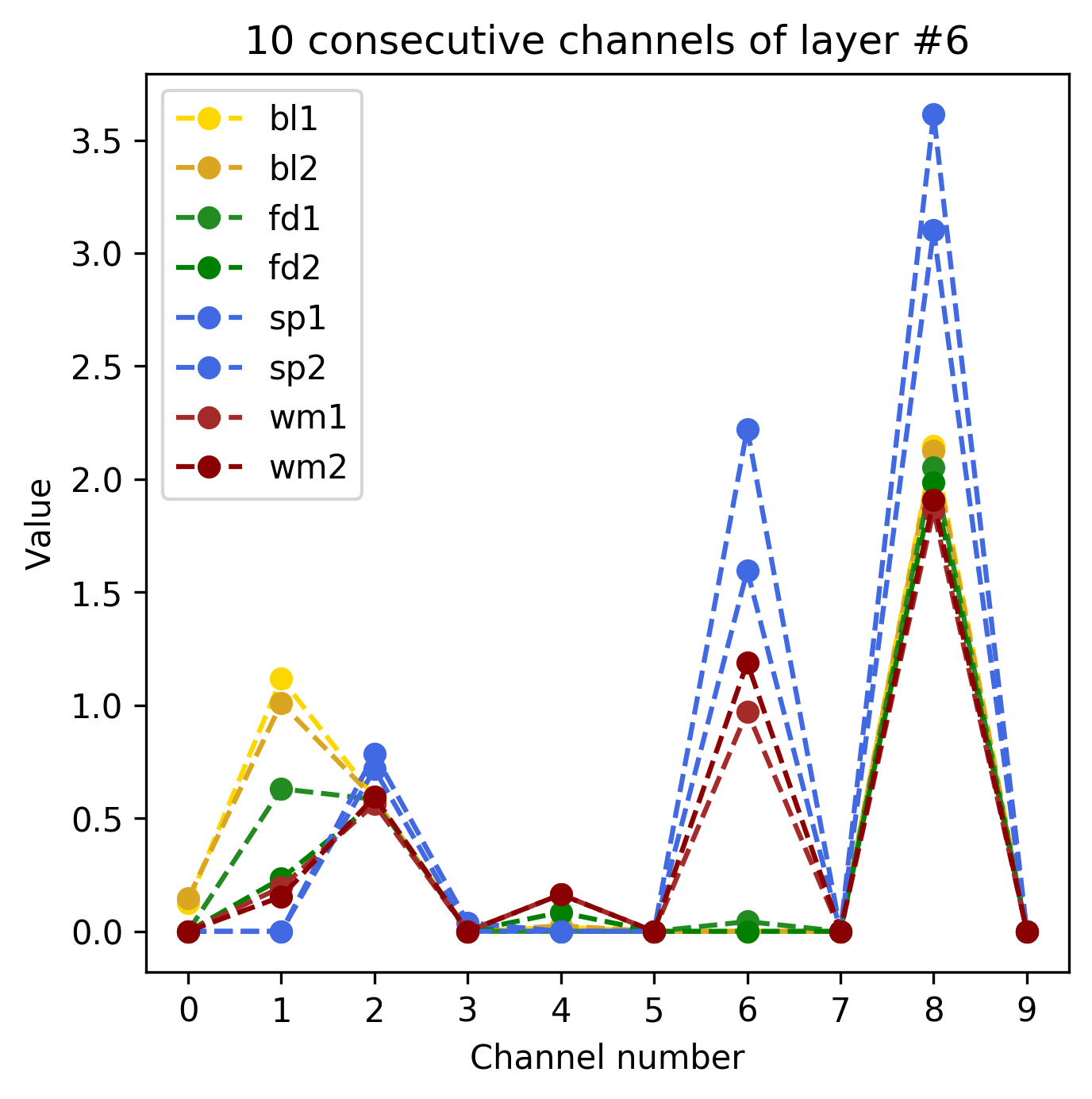}} 
    \subfloat[]{\label{fig:layer7} \includegraphics[width=.2\textwidth]{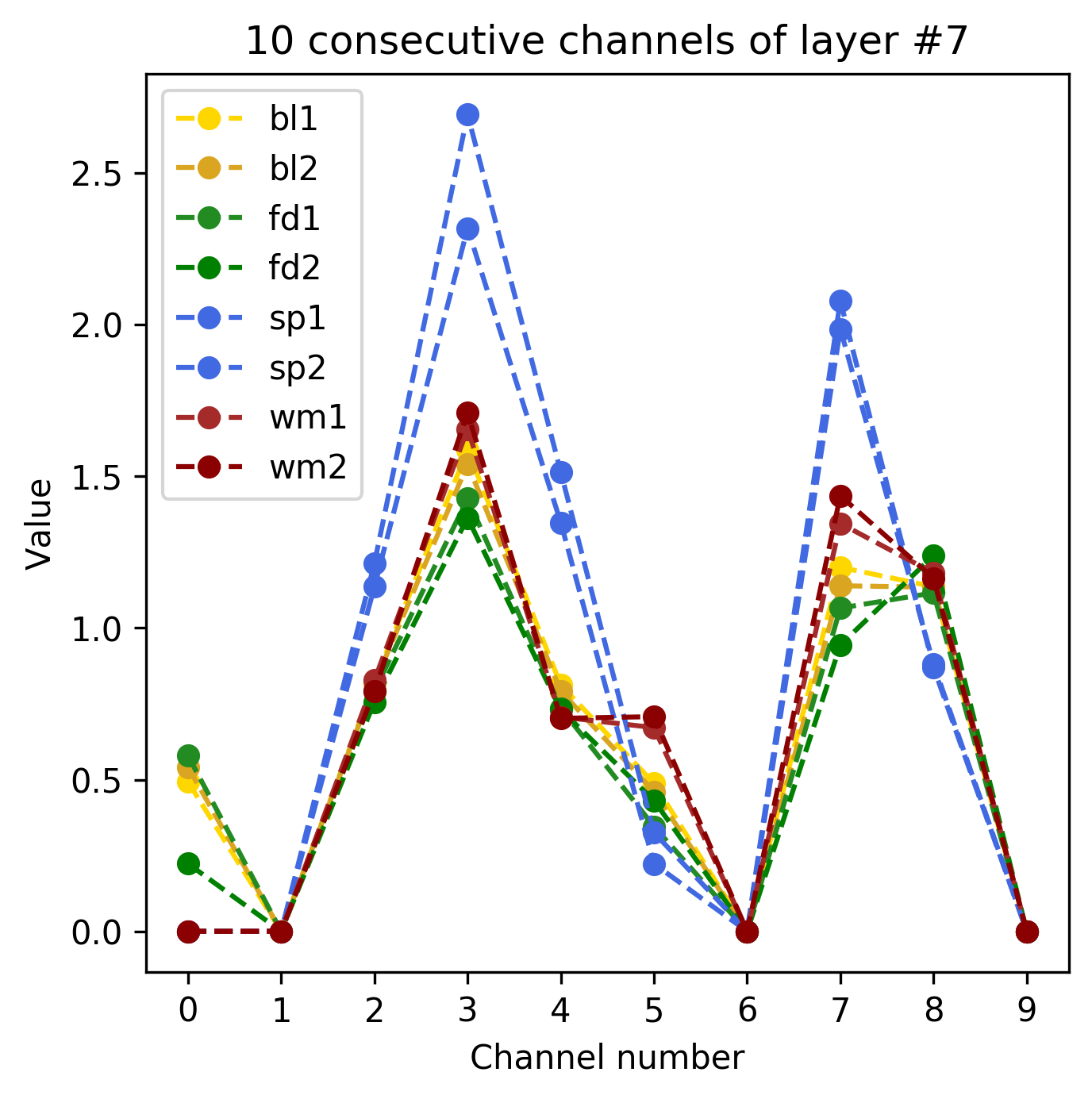}} 
    \subfloat[]{\label{fig:layer8} \includegraphics[width=.2\textwidth]{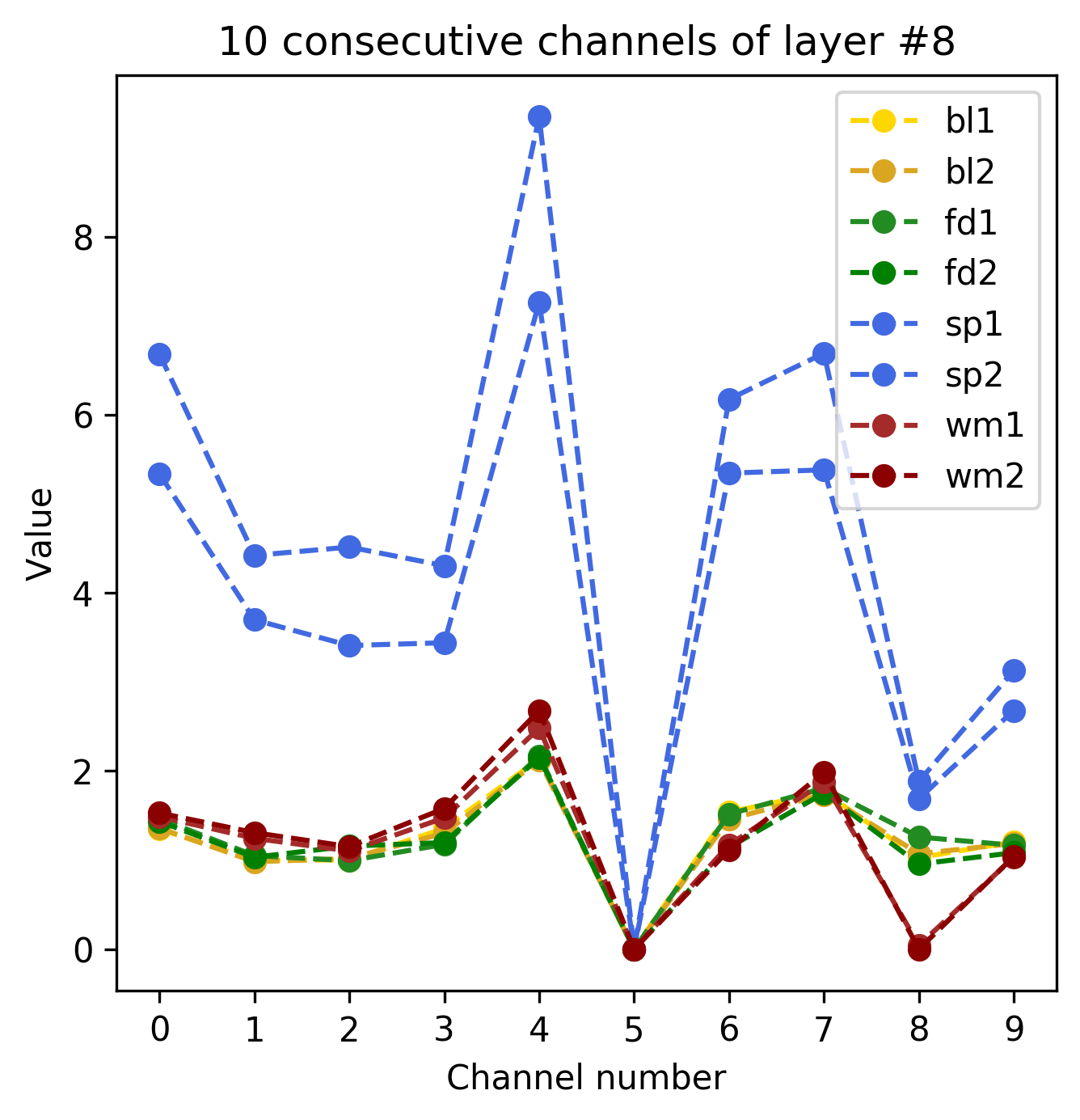}}
    \caption{Ten consecutive values of a section of gating network $g^*_H$ for all convolutional layers of forward generator (except layer three, for which the values are shown in Figure 2b of the main paper).}
    \label{fig:ablation_gate_resp}
\end{figure*}

In the main paper, we provided 10 consecutive values of a section of gating network $g^*_H$ for the third convolutional layer of forward generator. Here, in Figure~\ref{fig:ablation_gate_resp}, we provide these values for eight remaining convolutional layers of forward generator. These values were calculated for two samples of all considered noise types, including S\&P noise (blue), faded text (green), blurred text (yellow), and watermarked pages (red). These results are consistent with those displayed for layer three in Figure 2b of the main paper. As can be observed from these plots, there are strong similarities between the values generated by the gating networks for the same noise type, whereas they are different for different noise types. This demonstrates that the gating networks enable the forward generator to process an image in a different way depending on the containing noise type. 

\begin{figure*}[!tbh]
    \centering
    \subfloat[]{\label{fig:tsNE_gate0} \includegraphics[width=.2\textwidth]{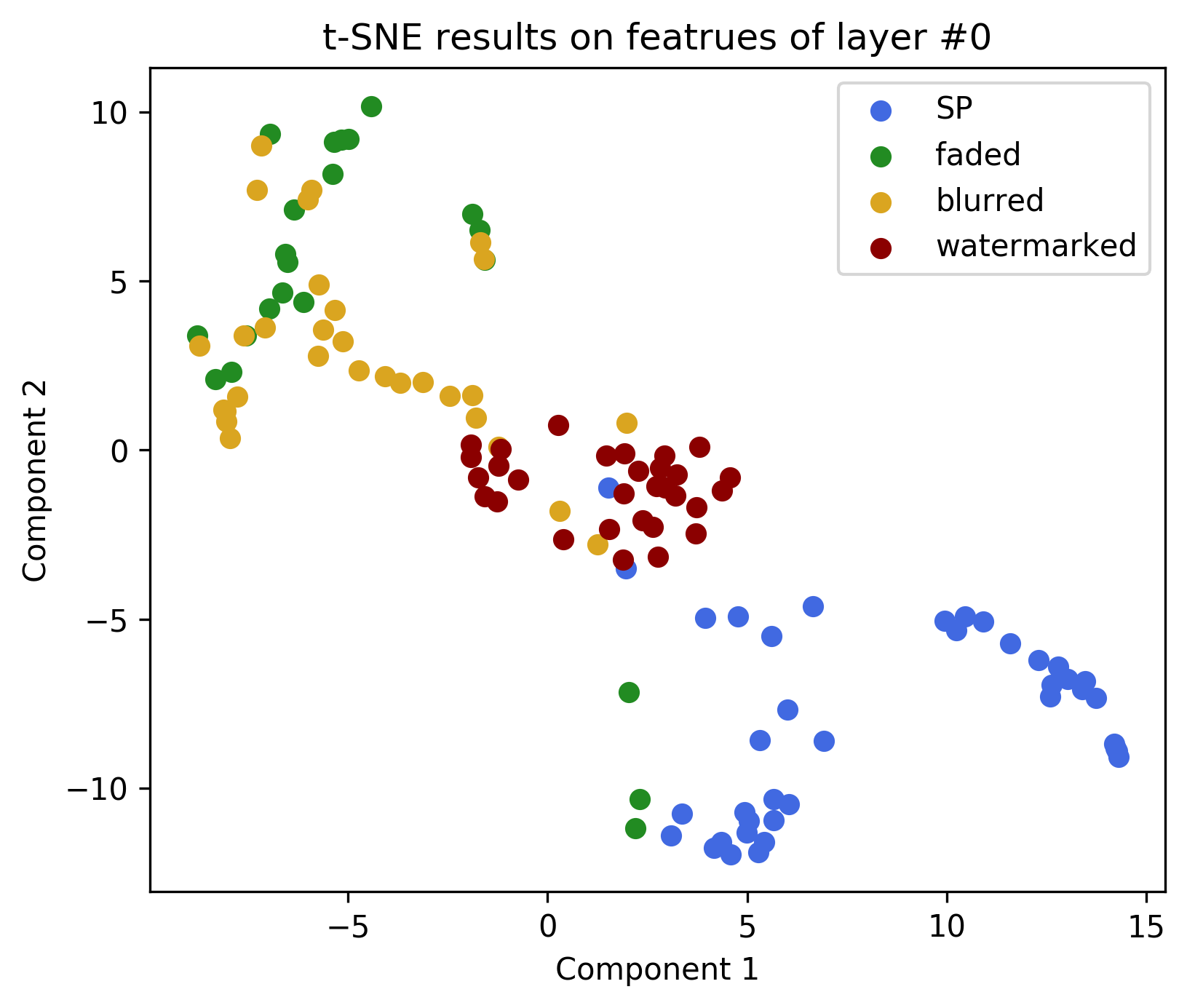}}
    \subfloat[]{\label{fig:tsNE_gate1} \includegraphics[width=.2\textwidth]{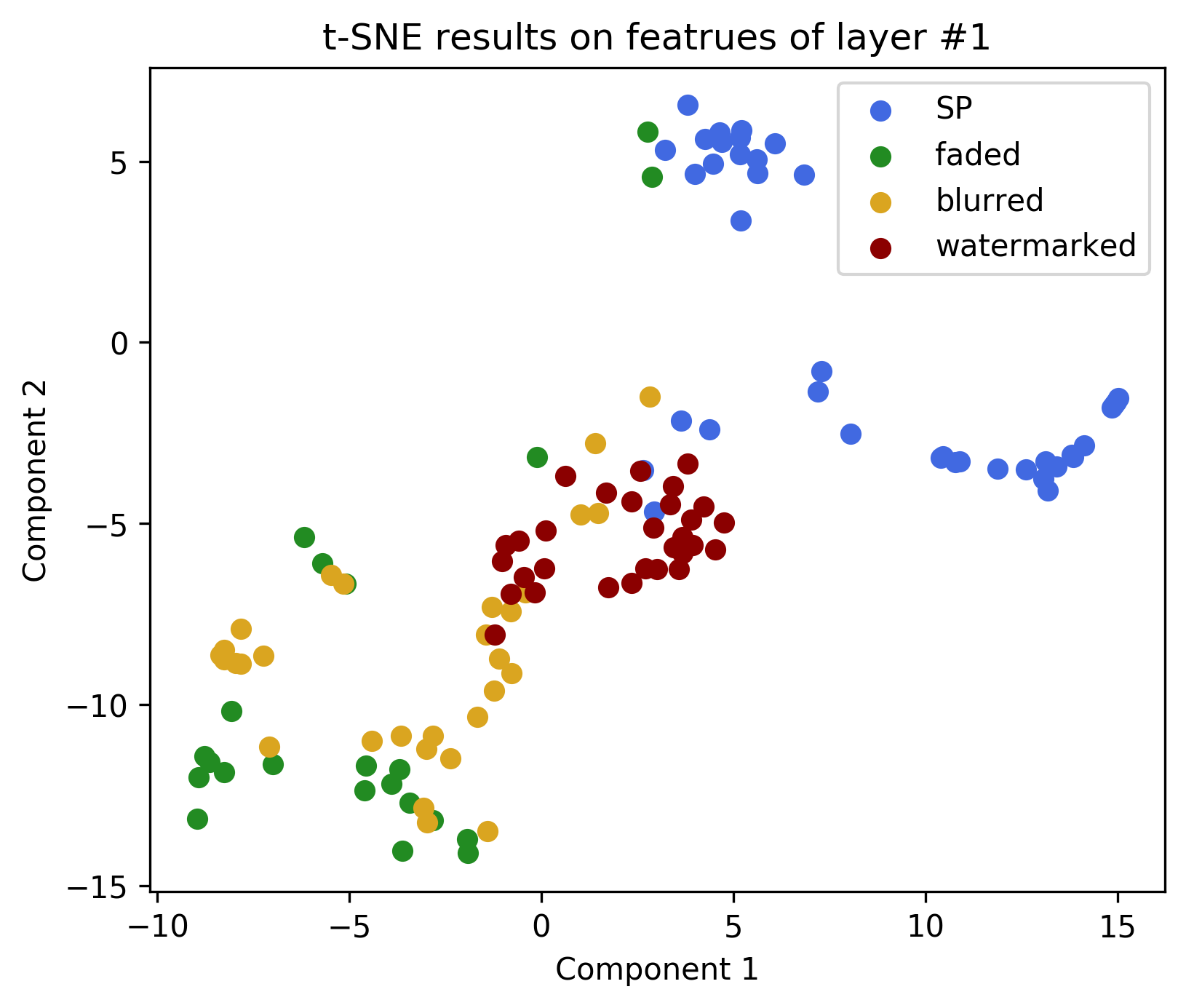}} 
    \subfloat[]{\label{fig:tsNE_gate2} \includegraphics[width=.2\textwidth]{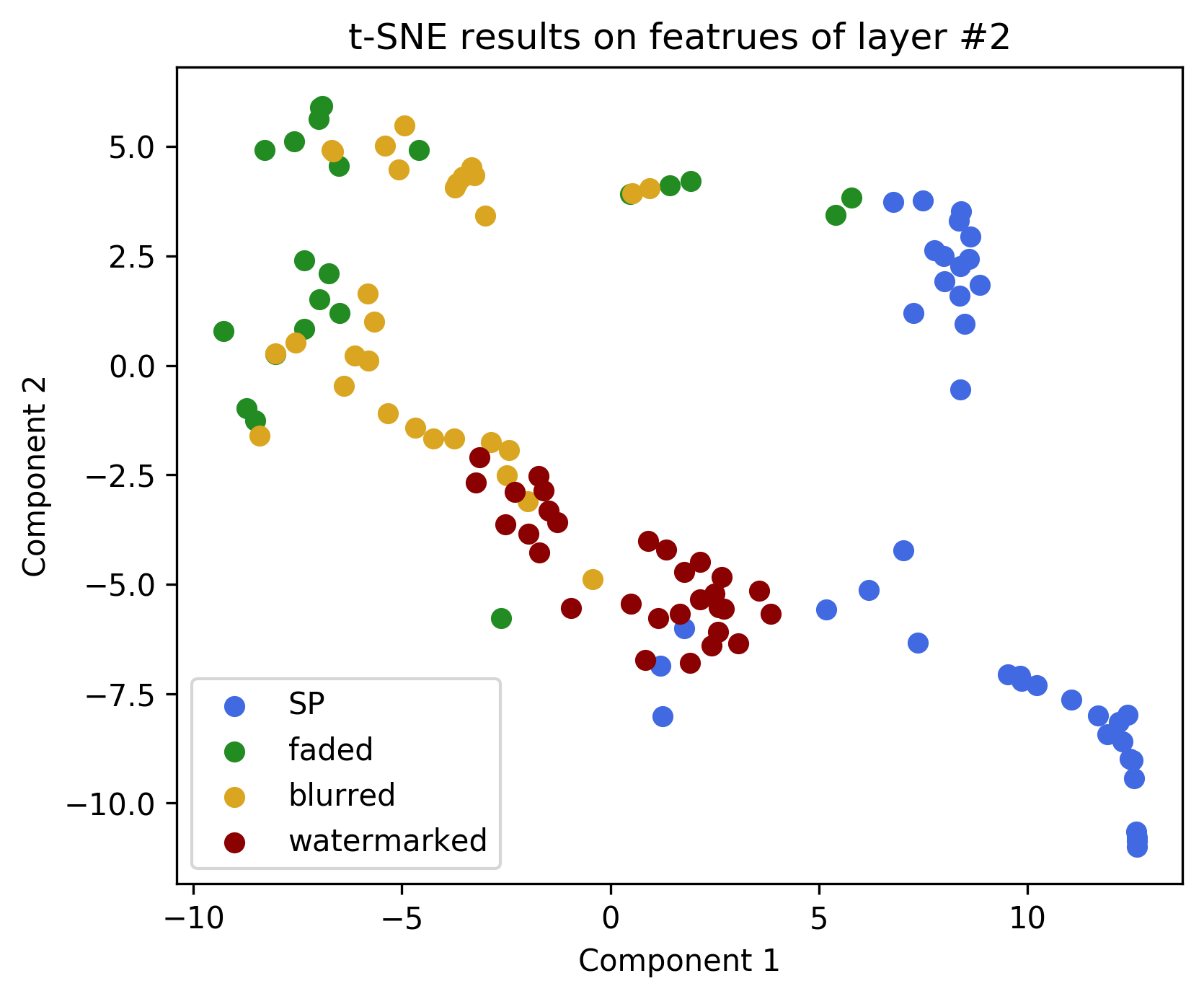}} \\ \vspace*{-0.9em}
    \subfloat[]{\label{fig:tsNE_gate3} \includegraphics[width=.2\textwidth]{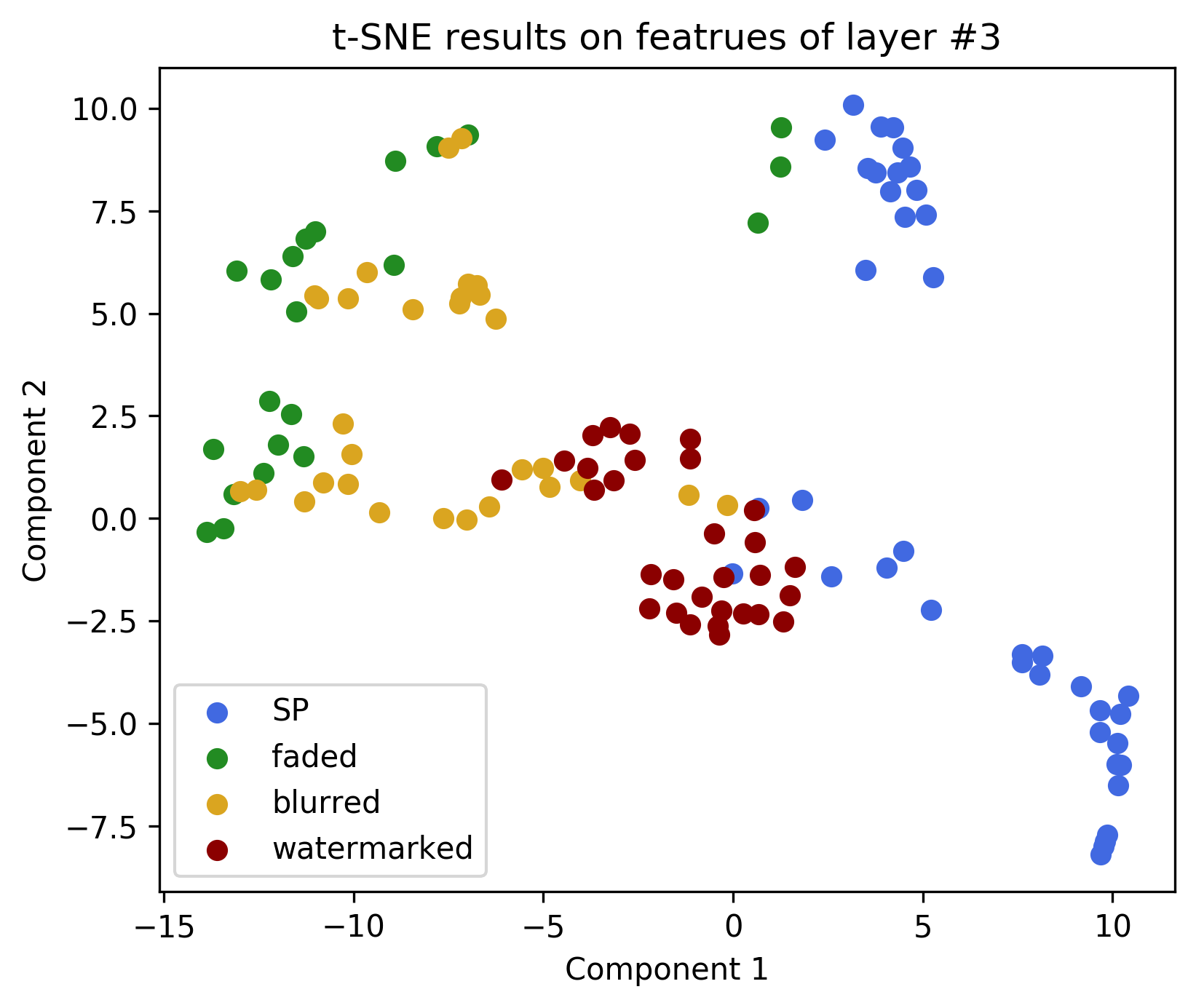}} 
    \subfloat[]{\label{fig:tsNE_gate4} \includegraphics[width=.2\textwidth]{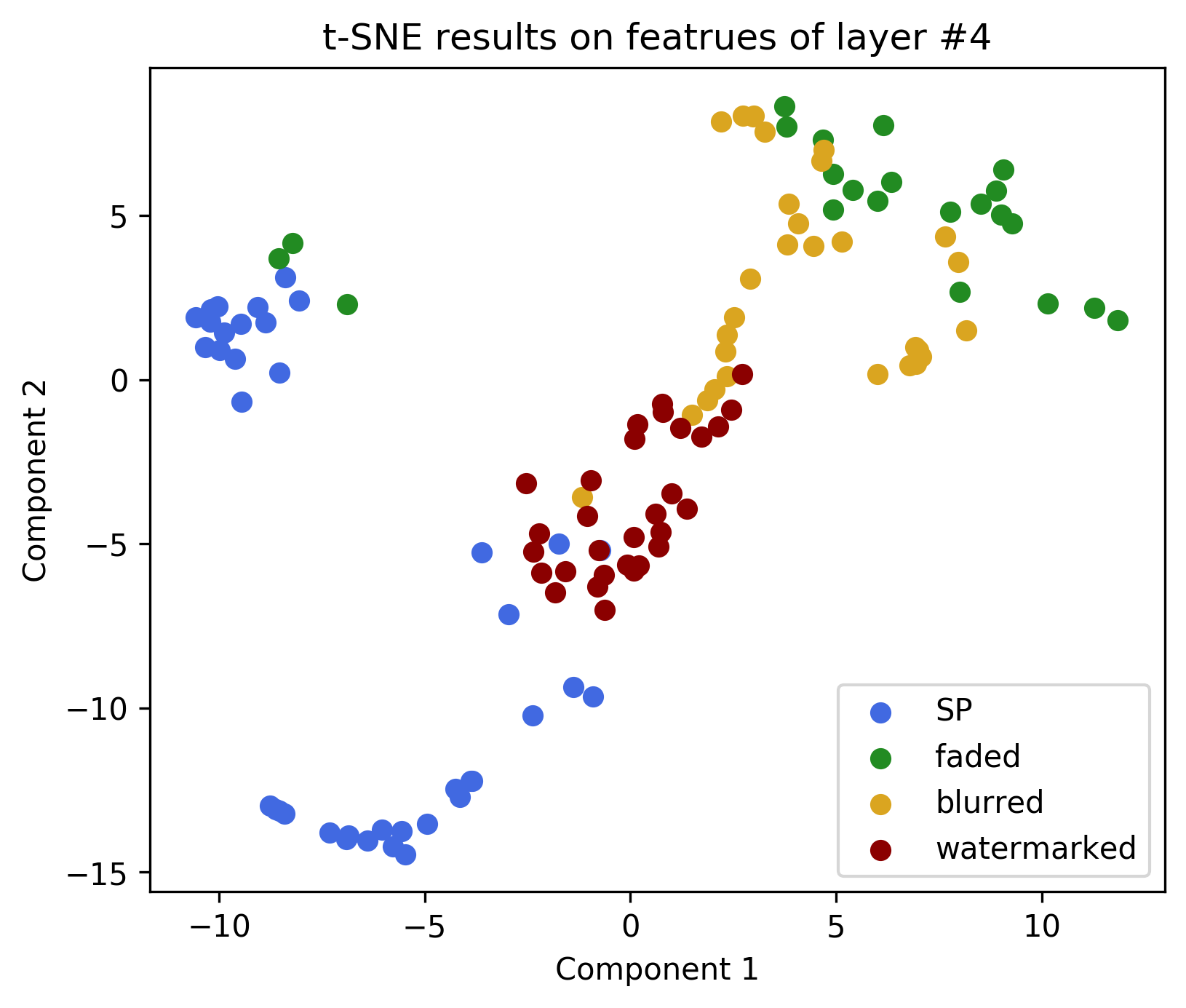}}
    \subfloat[]{\label{fig:tsNE_gate5} \includegraphics[width=.2\textwidth]{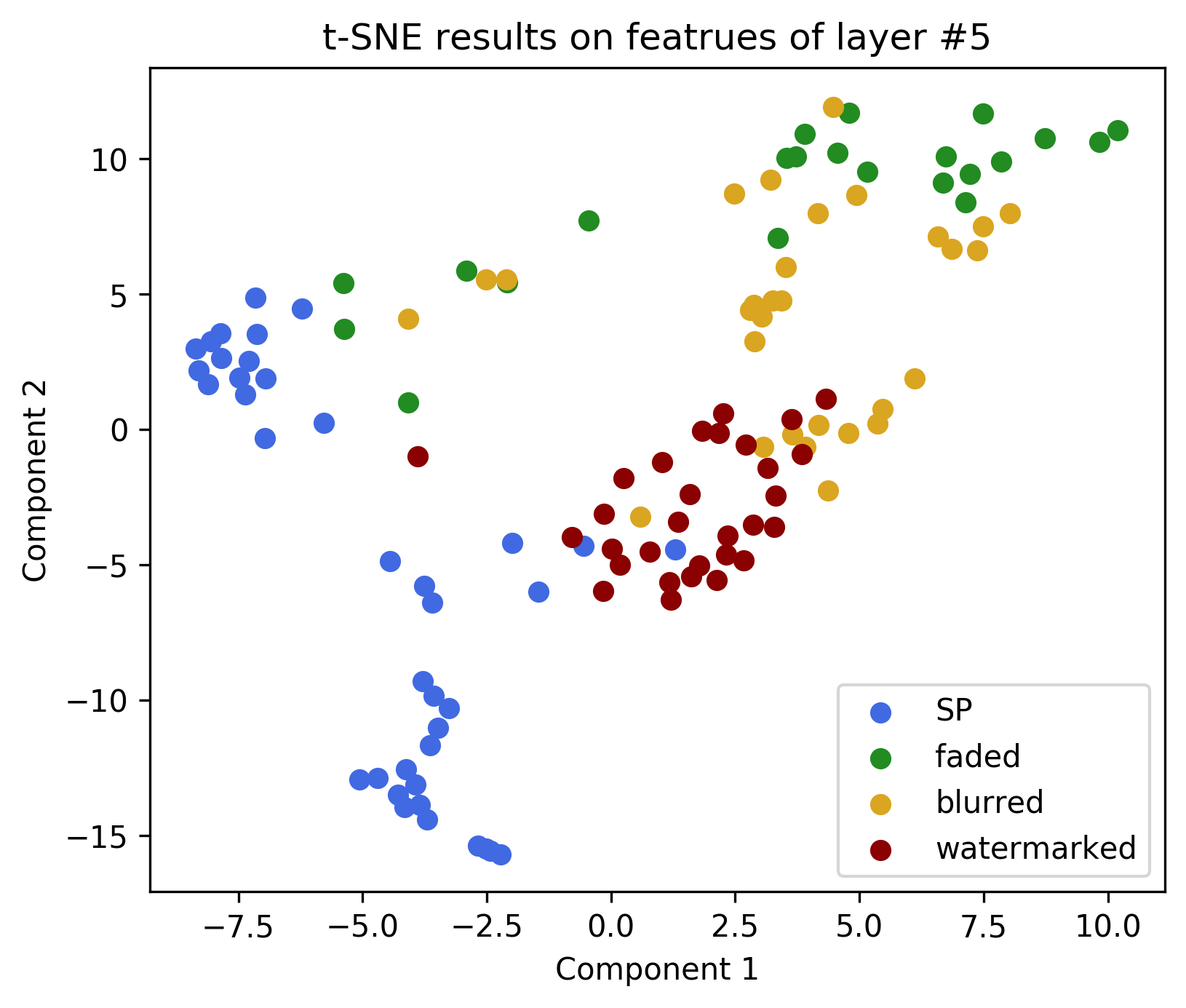}} \\ \vspace*{-0.9em}
    \subfloat[]{\label{fig:tsNE_gate6} \includegraphics[width=.2\textwidth]{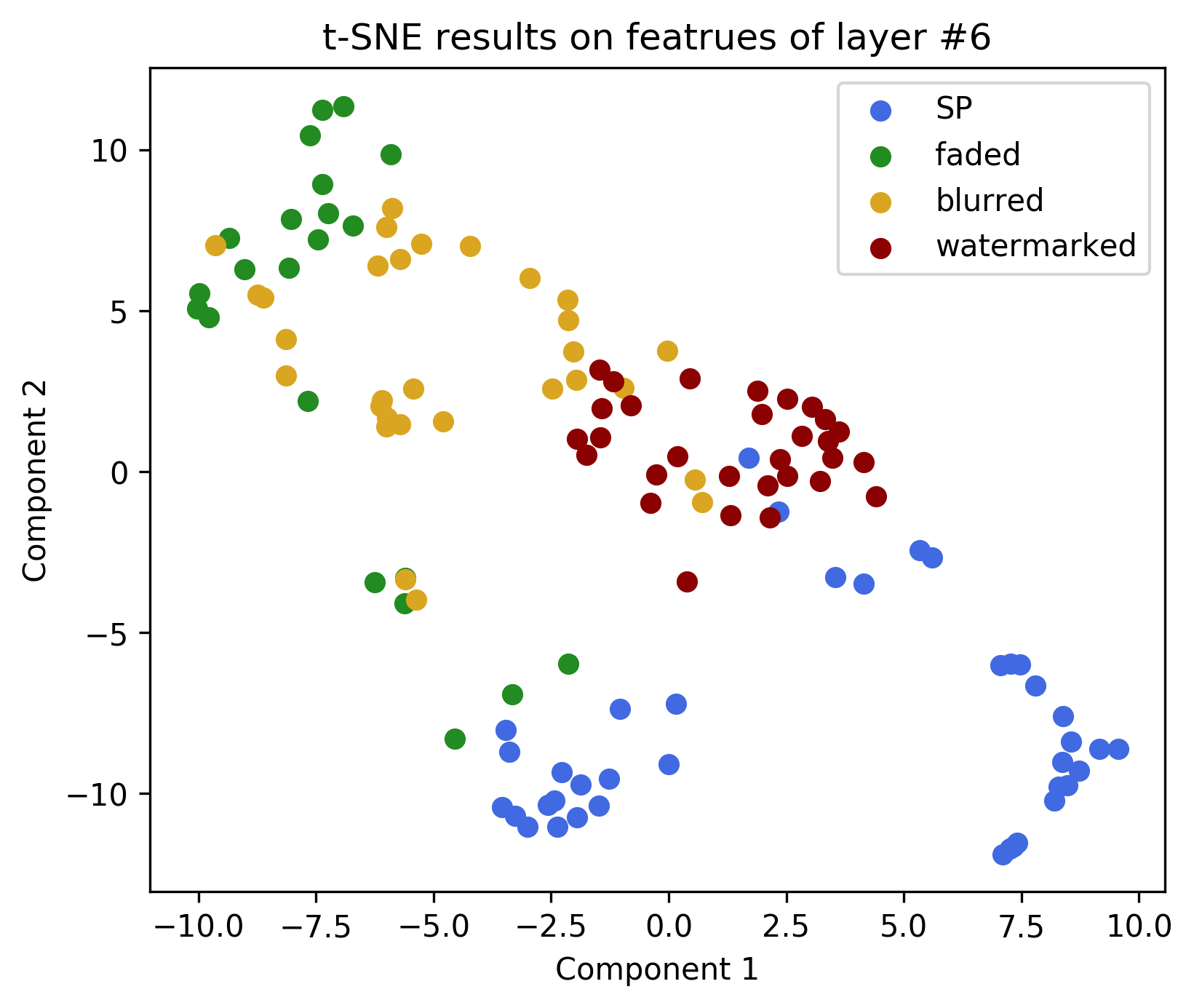}} 
    \subfloat[]{\label{fig:tsNE_gate7} \includegraphics[width=.2\textwidth]{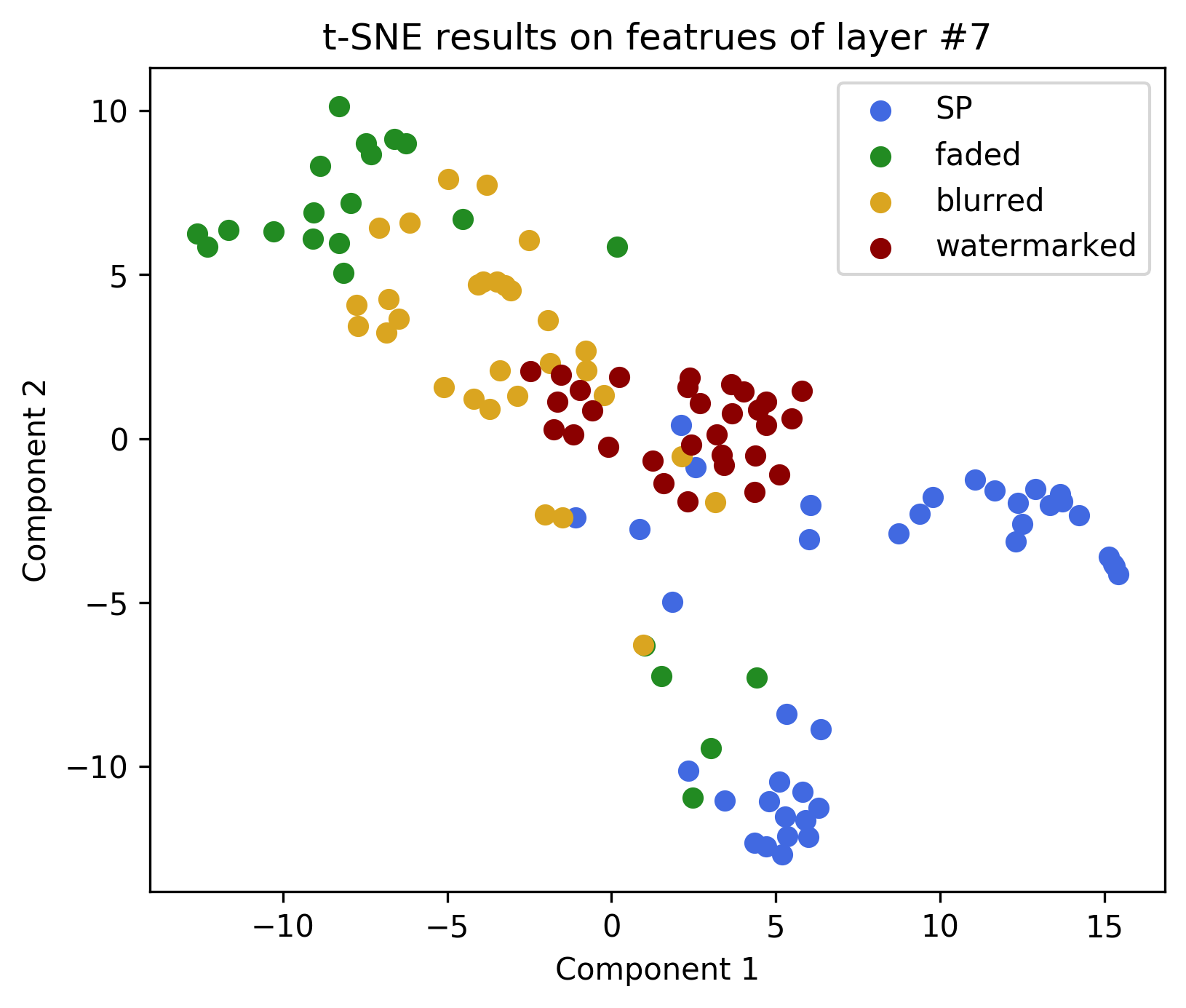}}
    \subfloat[]{\label{fig:tsNE_gate8} \includegraphics[width=.2\textwidth]{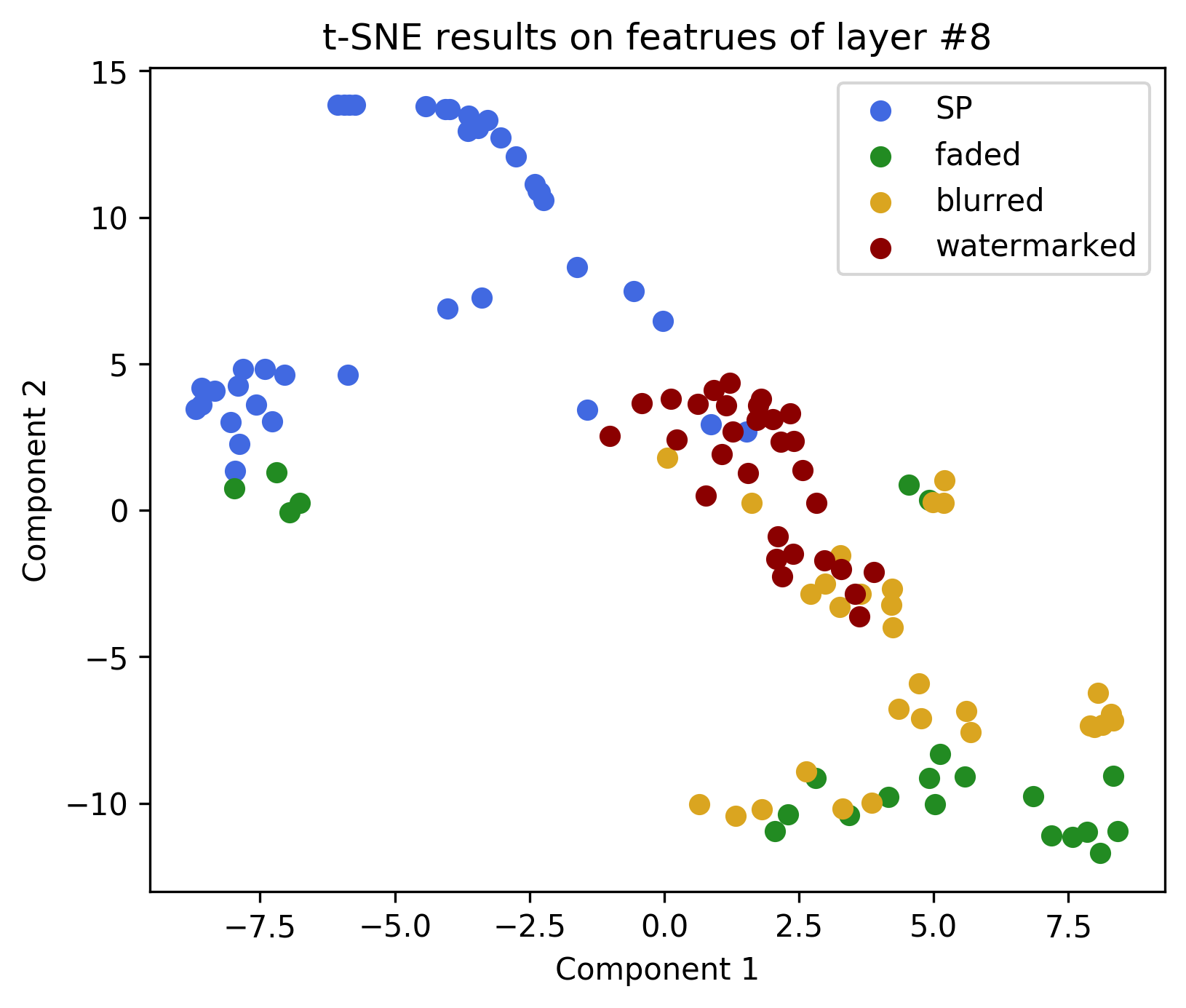}}
    \caption{ t-SNE~\cite{t-SNE} plots for all convolutional layers of forward generator provided for 120 document pages containing S\&P noise, faded or blurred text, or watermarks.}
    \label{fig:ablation_tSNE}
\end{figure*}

To further demonstrate the effectiveness of the gating networks, t-SNE~\cite{t-SNE} plots for all convolutional layers of forward generator are provided in Figure~\ref{fig:ablation_tSNE}. The plots depict the distributions of gate outputs (256 features) reduced to two main components using t-SNE algorithm. The plots are obtained for 120 document pages containing one of the artifact types, \ie, S\&P noise, faded or blurred text, or watermarks (30 pages in each category). The plots for all layers show that the gates outputs are well separated for all four artifact types, which demonstrates the ability of the gating networks to separate the various noise types. From the plots, it can be observed that the least characteristic features are related to blurred pages, as they are sometimes overlapped with the features calculated on faded or watermarked images. On the other hand, it can also be observed that pages containing S\&P noise make the most isolated class.

\subsection{Qualitative results}

\begin{figure*}[!tbh]
    \centering
    \subfloat[]{\label{fig:tobacco_sample1} \includegraphics[width=.98\textwidth]{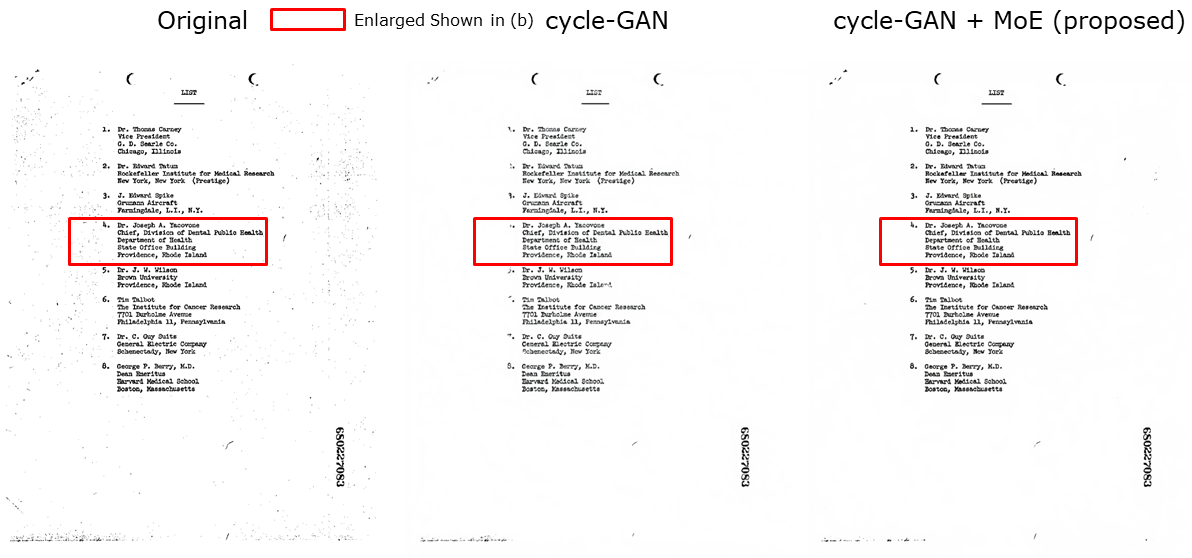}} \\ \vspace*{-0.9em}
    \subfloat[]{\label{fig:tobacco_patch11} \includegraphics[width=.98\textwidth]{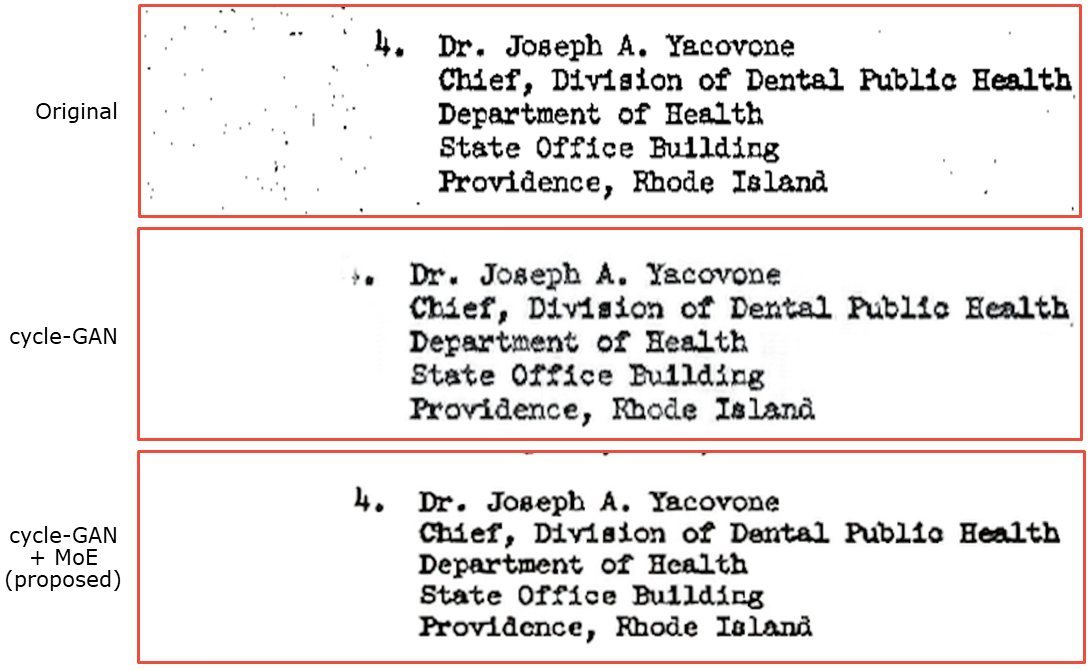}} 
    \caption{Qualitative results on a sample page from Tobacco800 Dataset (a) the whole page, (b) part of the page zoomed in (red box in (a)). The cleaned pages are compared between a standard cycle-GAN~\cite{Zhu2017} and the proposed approach.}
    \label{fig:tobacco}
\end{figure*}

\begin{figure*}[!tbh]
    \centering
    \subfloat[]{\label{fig:CDIP_sample1} \includegraphics[width=.9\textwidth]{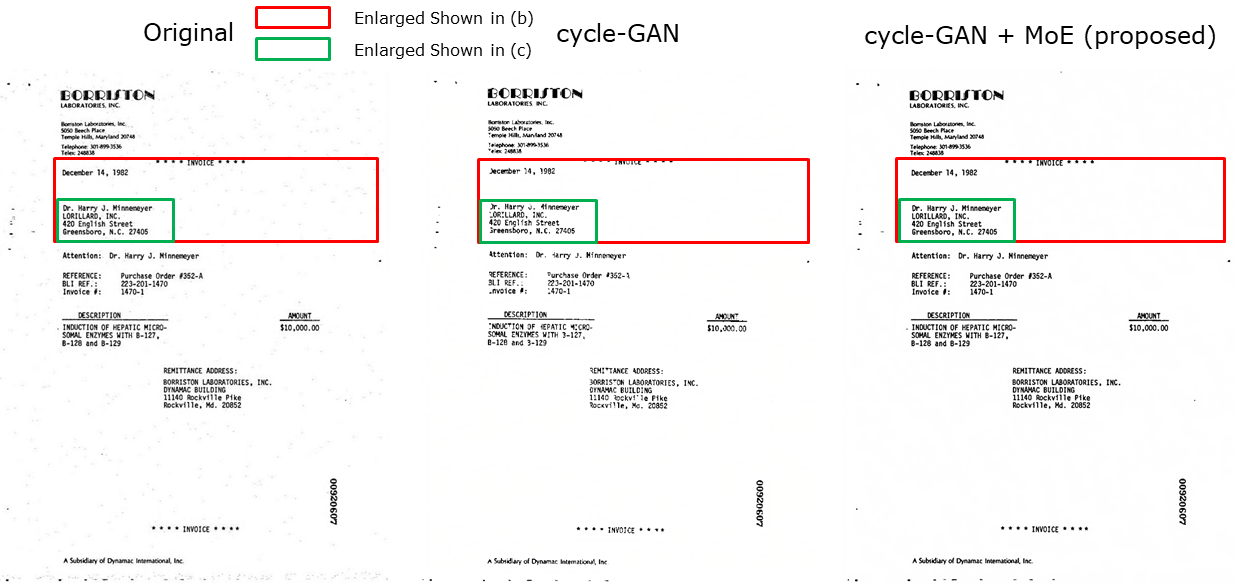}} \\ \vspace*{-0.9em}
    \subfloat[]{\label{fig:CDIP_patch11} \includegraphics[width=.7\textwidth]{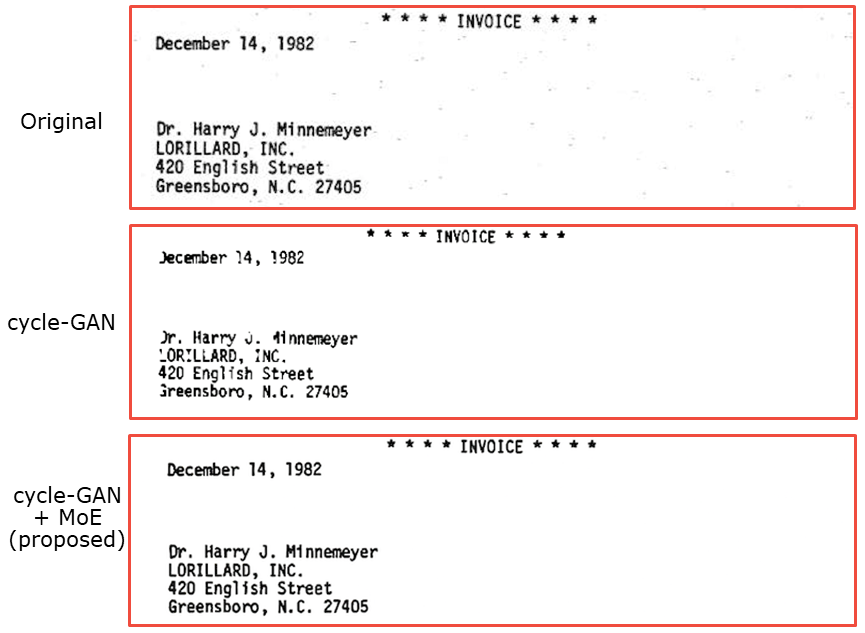}} \\ \vspace*{-0.9em}
    \subfloat[]{\label{fig:CDIP_patch12} \includegraphics[width=.99\textwidth]{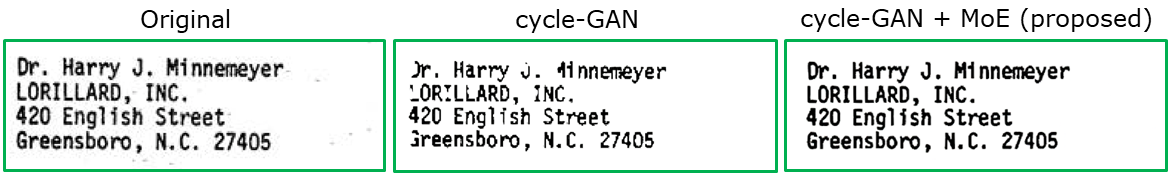}}
    \caption{Qualitative results on a sample page from CDIP Dataset (a) the whole page, (b) part of the page zoomed in (red box in (a)), (c) part of the page zoomed in (green box in (a)). The cleaned pages are compared between a standard cycle-GAN~\cite{Zhu2017} and the proposed approach.}
    \label{fig:CDIP1}
\end{figure*}

\begin{figure*}[!tbh]
    \centering
    \subfloat[]{\label{fig:CDIP_sample2} \includegraphics[width=.9\textwidth]{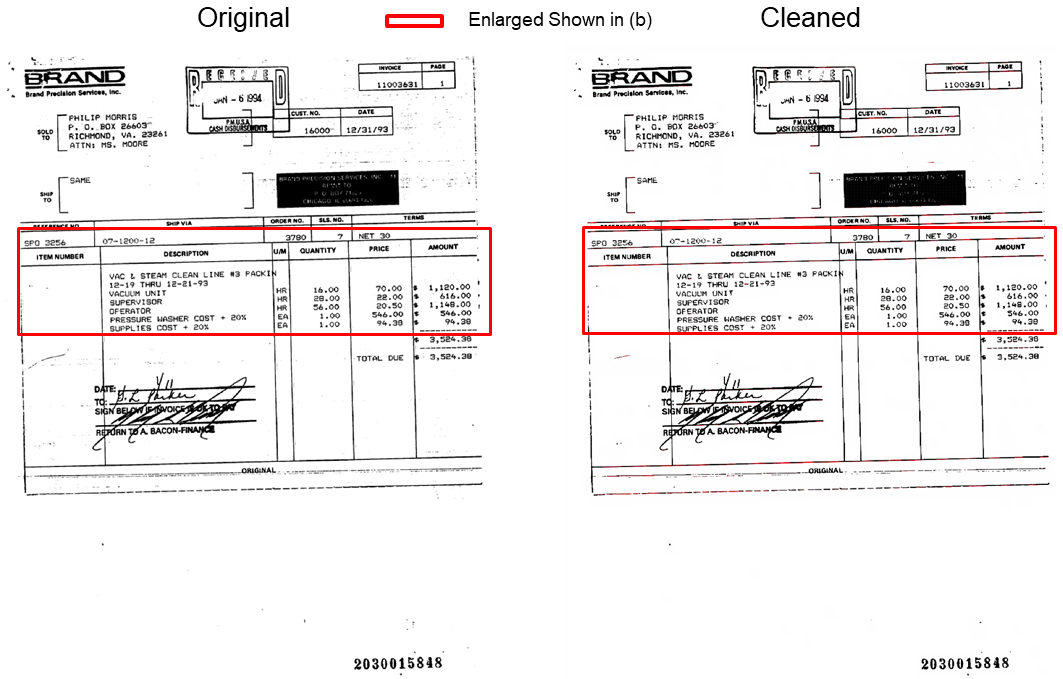}} \\ 
    \subfloat[]{\label{fig:CDIP_patch21} \includegraphics[width=.9\textwidth]{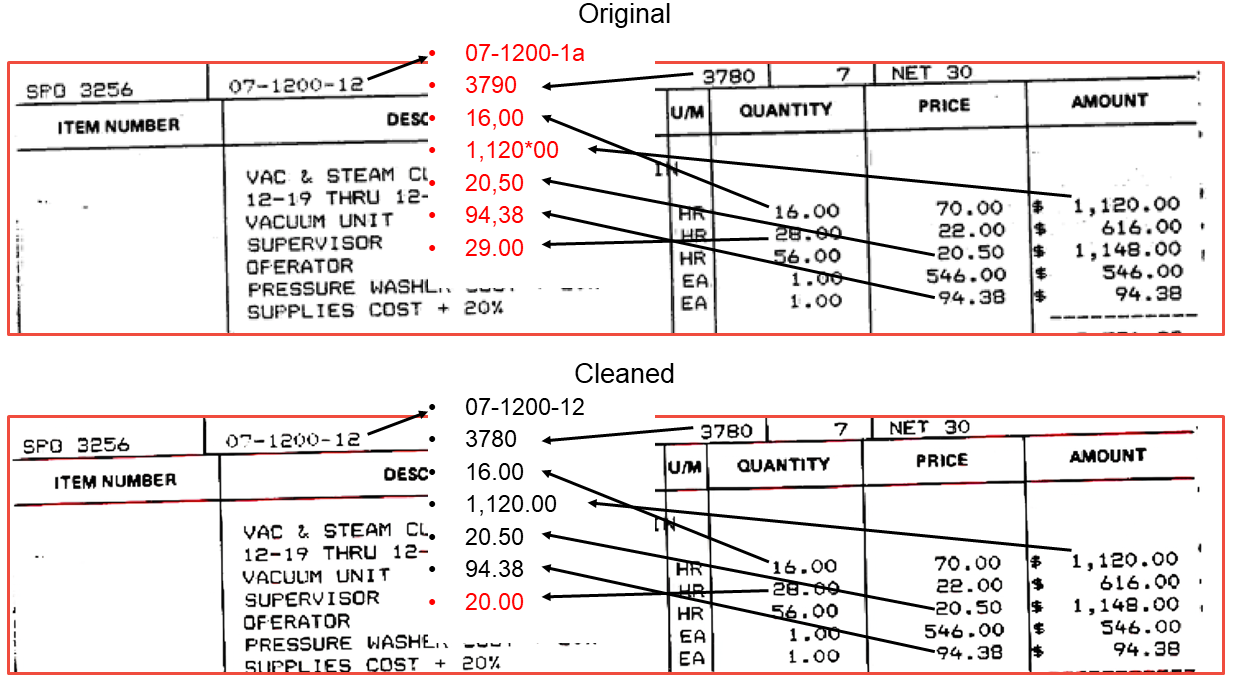}} \\ 
    \caption{Qualitative results on a sample page from CDIP Dataset (a) the whole page, (b) part of the page zoomed in (red box in (a)), the words that result in different OCR on the original and cleaned pages are also displayed for comparison.}
    \label{fig:CDIP2}
\end{figure*}

\begin{figure*}[!tbh]
\begin{center}
  \includegraphics[width=0.95\linewidth]{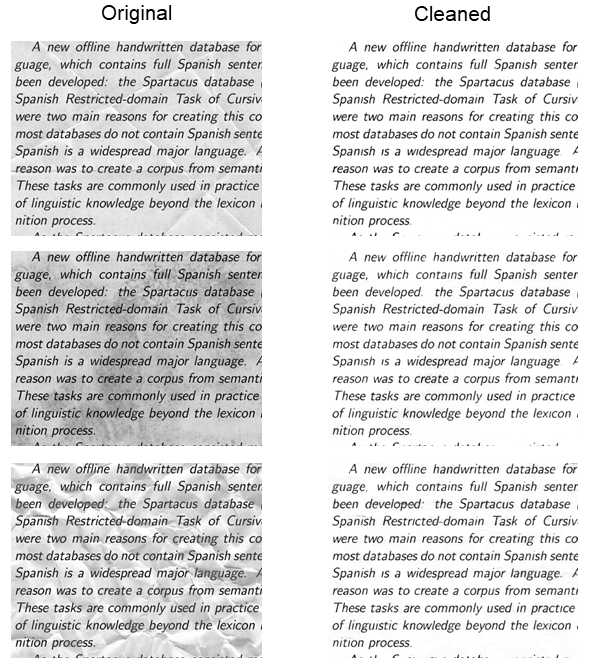}
\end{center}
  \caption{Qualitative results on a few samples from Kaggle Dataset.} 
\label{fig:Kaggle}
\end{figure*}

\begin{figure*}[!tbh]
    \centering
    \subfloat[]{\label{fig:watermarked-K1} \includegraphics[width=.33\textwidth]{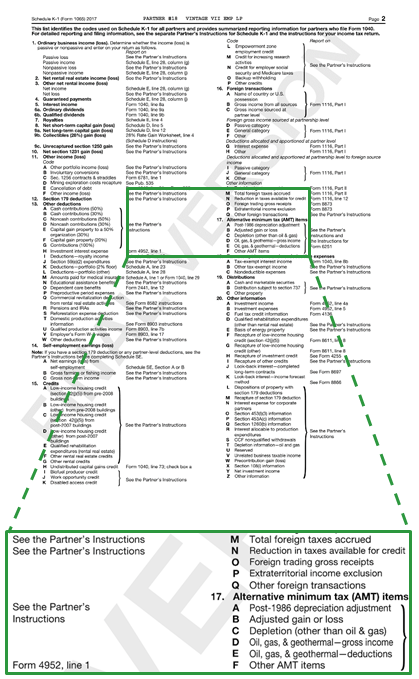}} 
    \subfloat[]{\label{fig:watermarked_cleaned_RED-Net-K1} \includegraphics[width=.33\textwidth]{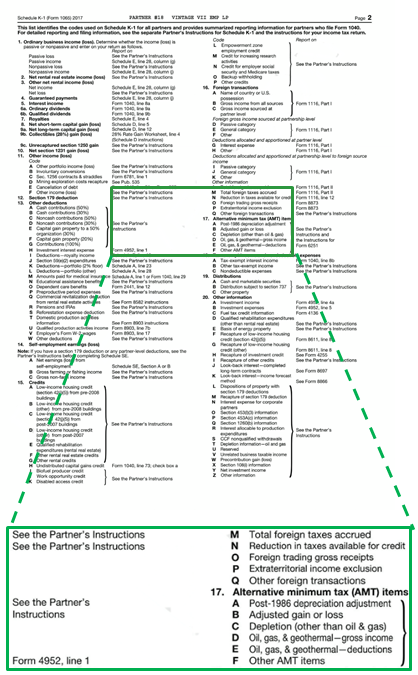}} \\ 
    \subfloat[]{\label{fig:watermarked_cleaned_DE-GAN-K1} \includegraphics[width=.33\textwidth]{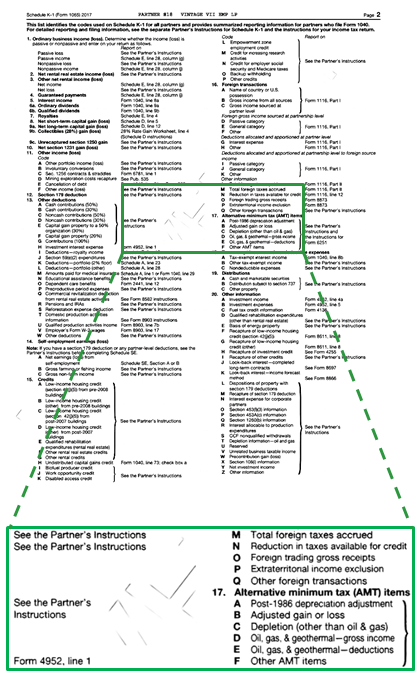}} 
    \subfloat[]{\label{fig:watermarked_cleaned_moe-K1} \includegraphics[width=.33\textwidth]{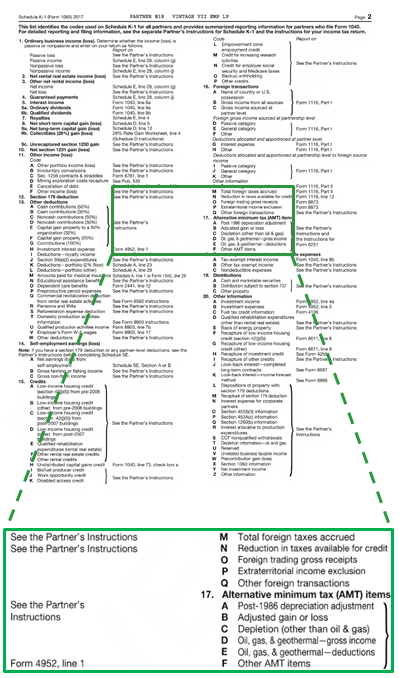}} 
    \caption{Qualitative results on a sample instruction page of a tax form: (a) watermarked page, and cleaned pages using: (b) RED-Net~\cite{RedNet2016}, (c) DE-GAN~\cite{DE-GAN2020}, and (d) proposed approach.}
    \label{fig:watermarked_K1}
\end{figure*}

\begin{figure*}[!tbh]
    \centering
    \subfloat[]{\label{fig:watermarked-46} \includegraphics[width=.28\textwidth]{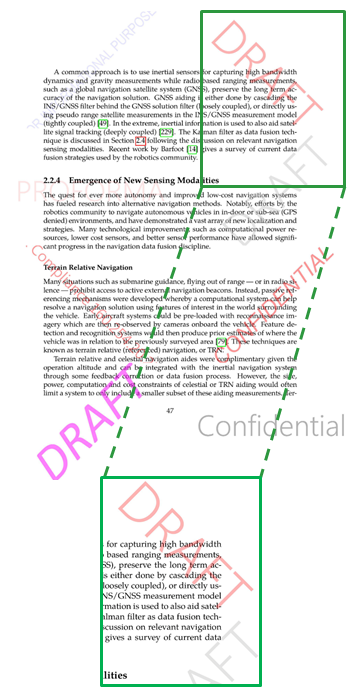}} 
    \subfloat[]{\label{fig:watermarked_cleaned_RED-Net-46} \includegraphics[width=.28\textwidth]{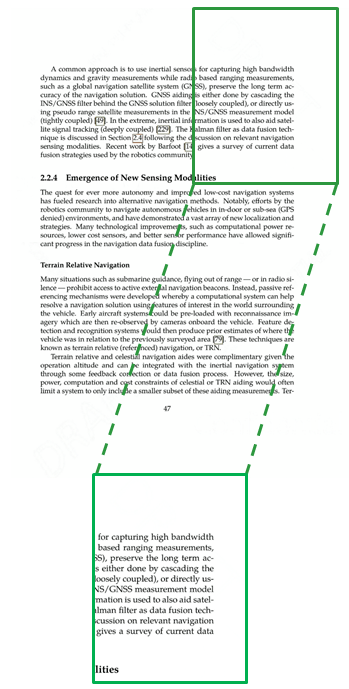}} \\ 
    \subfloat[]{\label{fig:watermarked_cleaned_DE-GAN-46} \includegraphics[width=.28\textwidth]{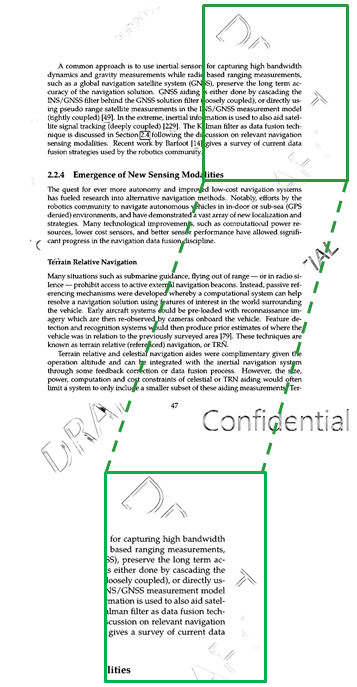}} 
    \subfloat[]{\label{fig:watermarked_cleaned_moe-46} \includegraphics[width=.28\textwidth]{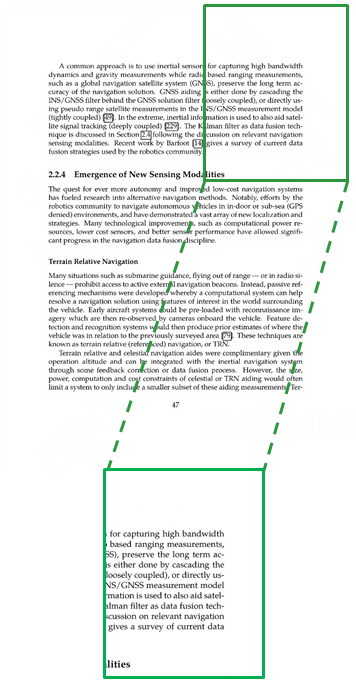}} 
    \caption{Qualitative results on a scientific paper with synthetically added watermarks: (a) watermarked page, and cleaned pages using: (b) RED-Net~\cite{RedNet2016}, (c) DE-GAN~\cite{DE-GAN2020}, and (d) proposed approach.}
    \label{fig:watermarked_46}
\end{figure*}

\begin{figure*}[!tbh]
    \centering
    \subfloat[]{\label{fig:watermarked-49} \includegraphics[width=.28\textwidth]{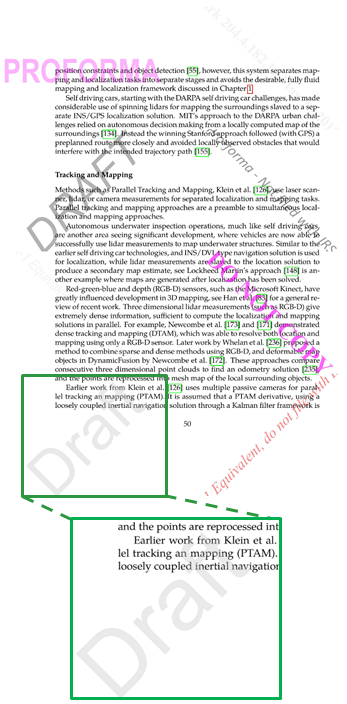}} 
    \subfloat[]{\label{fig:watermarked_cleaned_RED-Net-49} \includegraphics[width=.28\textwidth]{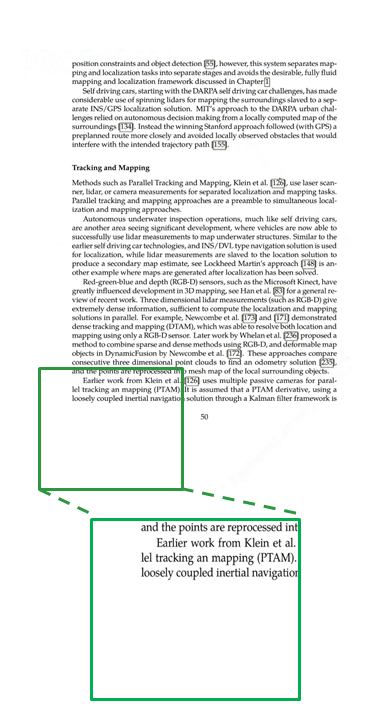}} \\ 
    \subfloat[]{\label{fig:watermarked_cleaned_DE-GAN-49} \includegraphics[width=.28\textwidth]{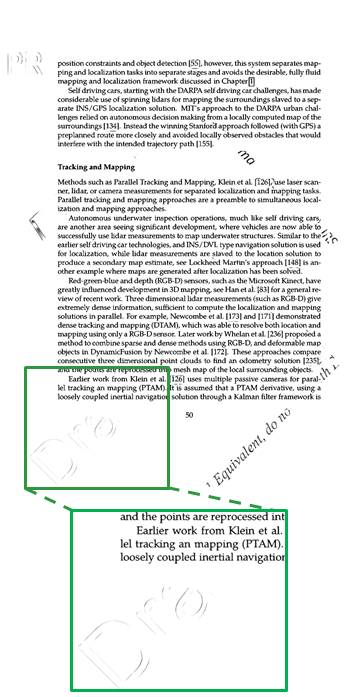}} 
    \subfloat[]{\label{fig:watermarked_cleaned_moe-49} \includegraphics[width=.28\textwidth]{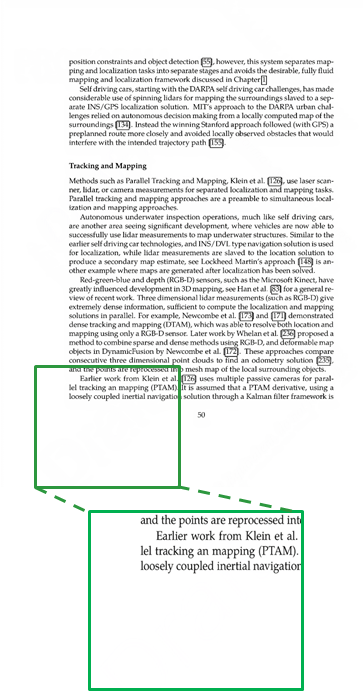}} 
    \caption{Qualitative results on a scientific paper with synthetically added watermarks: (a) watermarked page, and cleaned pages using: (b) RED-Net~\cite{RedNet2016}, (c) DE-GAN~\cite{DE-GAN2020}, and (d) proposed approach.}
    \label{fig:watermarked_49}
\end{figure*}

We have provided more results on a few noisy document pages, including various artifacts, such as S\&P noise, faded or blurred text, and watermarks. These results are provided in  Figure~\ref{fig:tobacco} for a page from Tobacco800 dataset~\cite{SignatureDetection-CVPR07}, Figures~\ref{fig:CDIP1} and~\ref{fig:CDIP2} for two sample pages from CDIP dataset~\cite{CDIP_dataset}, Figure~\ref{fig:Kaggle} for a few samples from Kaggle dataset~\cite{Kaggle_dataset}, and Figures~\ref{fig:watermarked_K1}, \ref{fig:watermarked_46}, and \ref{fig:watermarked_49} for an instruction page of a tax form and two pages from a scientific paper with synthetically added watermarks, respectively. 
In order to demonstrate the effectiveness of the proposed approach in image clean-up, we have compared our approach with the standard cycle-GAN~\cite{Zhu2017} (without integrated deep MoE) trained on multiple noise types, including S\&P noise, faded, and blurred pages in Figures~\ref{fig:tobacco} and \ref{fig:CDIP1}. As can be observed from these two figures, the proposed model is much more effective in removing noise without distorting the texts on the pages. Furthermore, in order to demonstrate the improvement in the OCR after removing noise from a page, we have depicted the differences in OCR on part of a page before and after cleansing in Figure~\ref{fig:CDIP_patch21}. It can be observed that cleansing the page using the proposed approach is quite effective to improve the OCR performance and to generate correct OCR on the cleaned page. It is important to note that the proposed model has not been trained on any samples from Tobacco800, Kaggle, or CDIP datasets. As was explained in Section~\ref{sec:training}, the model has only been trained on our in-house documents, including lease contracts, tax forms, and invoices. Nonetheless, it produces excellent noise removal performance across these public datasets as demonstrated in the results depicted on this supplementary note. 

For watermark removal problem, we have qualitatively compared the proposed approach with two supervised approaches, \ie, REDNet~\cite{RedNet2016} and DE-GAN~\cite{DE-GAN2020} in Figures~\ref{fig:watermarked_K1}, \ref{fig:watermarked_46}, and \ref{fig:watermarked_49}. REDNet and DE-GAN have solely been trained using part of our training dataset containing paired watermarked/clean patches extracted from tax forms, whereas our proposed method has been trained on all noise types (S\&P, faded, blurred, and watermarked). As can be observed in Figures~\ref{fig:watermarked_cleaned_DE-GAN-K1}, \ref{fig:watermarked_cleaned_DE-GAN-46}, and \ref{fig:watermarked_cleaned_DE-GAN-49}, DE-GAN has difficulty to remove watermark from blank parts of the pages and this should be due to the additional loss function the authors have introduced to the model to preserve the text on pages (refer to Eq. (2) and corresponding explanations in~\cite{DE-GAN2020}). Although our proposed model is unsupervised and has been trained for several noise types, it is as effective as REDNet (a supervised approach solely trained for watermark removal) in removing watermark from pages. 


\end{document}